\documentclass[twoside,11pt]{article}

\usepackage{jmlr2e}

\usepackage{algorithm} 
\usepackage[noend]{algpseudocode} 
\usepackage{amsmath}
\usepackage{array}
\usepackage{tabularx}
\usepackage{multirow}

\ShortHeadings{Automated Machine Learning for Positive-Unlabelled Learning}{Saunders and Freitas}
\firstpageno{1}

\begin{document}

\title{Automated Machine Learning \\for Positive-Unlabelled Learning}

\author{\name Jack D.\ Saunders \email jsaunders0027@gmail.com \\
       \addr School of Computing,
       University of Kent\\
       Canterbury, CT2 7NT, UK
       \AND
       \name Alex A.\ Freitas \email A.A.Freitas@kent.ac.uk \\
       \addr School of Computing,
       University of Kent\\
       Canterbury, CT2 7NT, UK}

\editor{Editor Name}

\maketitle

\begin{abstract}
Positive-Unlabelled (PU) learning is a growing field of machine learning
that aims to learn classifiers from data consisting of labelled positive
and unlabelled instances, which can be in reality positive or negative, but 
whose label is unknown. An extensive number of methods have been proposed
to address PU learning over the last two decades, so many so that selecting an optimal
method for a given PU learning task presents a challenge. Our previous work
has addressed this by proposing GA-Auto-PU, the first Automated Machine 
Learning (Auto-ML) system for PU learning. In this work, we propose two new Auto-ML systems for PU learning: 
BO-Auto-PU, based on a Bayesian Optimisation approach, and EBO-Auto-PU, based on a novel evolutionary/Bayesian optimisation approach. We also present an extensive evaluation of the three Auto-ML systems, comparing them to each other 
and to well-established PU learning methods across 60 datasets (20 real-world datasets, each with 3 versions in terms of PU learning characteristics).
\end{abstract}


\section{Introduction}
Positive-Unlabelled (PU) learning is a field of machine learning that aims to learn classifiers from data consisting of a set of labelled positive instances and a set of unlabelled instances which can be either positive or negative, but whose label is unknown \citep{bekker2020learning}. To accurately utilise a standard machine learning method, a fully labelled dataset is required. However, in real-world applications this is often unfeasible due to the expense or impracticality of obtaining fully labelled data \citep{bekker2020learning}. To address this issue, many methods have been proposed to learn classifiers from partially labelled data \citep{bekker2020learning} \citep{jaskie2019positive}. This field is referred to as semi-supervised classification and usually focuses on learning classifiers from data where only a subset of instances is labelled as the positive or negative class, whilst all the other instances (usually the majority of instances) are unlabelled. PU learning is a specialised case of semi-supervised learning where, among the truly positive-class instances, only a subset is labelled as positive; whilst all the other instances are unlabelled – i.e., it is unclear if they are positive or negative instances \citep{bekker2020learning}. Thus, positive-unlabelled learning presents an arguably greater challenge than the standard semi-supervised learning paradigm, due to the complete absence of negative labels in the data. 

There are many real world applications of PU learning, including cybersecurity \citep{zhang2017pu} \citep{luo2018pu}, bioinformatics \citep{yang2012positive-unlabelled} \citep{nikdelfaz2018disease} \citep{vasighizaker2018c}, and text mining \citep{liu2002partially} \citep{ke2012building} \citep{liu2014clustering}. For example, PU learning has previously been used for predicting disease-related genes  \citep{nikdelfaz2018disease}. This is a PU learning task where disease-associated genes are the labelled positive instances, as confirmed by biomedical experiments. However, the vast majority of the genes thought not to be associated with diseases have not undergone such experiments, since these experiments are expensive. As such, the genes without association with diseases are better thought of as unlabelled instances as there is no experimental evidence indicating either association or disassociation. An example from the domain of text mining is found in \citep{liu2014clustering}, which proposed a text classification system utilising PU learning for web page classification. This is another learning task where PU learning is appropriate. Scraping web pages is an easy and quick task, so assembling a large dataset is a simple process. However, the majority of the instances (webpages) will be unlabelled as manually labelling each instance is an expensive task. As illustrated by these examples, PU learning is appropriate when the dataset consists of a small sample of reliable positives and a larger remaining sample of unknown-label instances.

The most popular approach to PU learning is the two-step approach \citep{bekker2020learning}, which aims, in step 1, to find a set of reliable negative instances in the unlabelled set. That is, a set of instances which are substantially different from the positive set and are likely not unlabelled positive instances. In step 2, a classifier is trained to distinguish between the labelled positive instances and the reliable negative instances, building a classifier which, if the reliable negative set is accurate, can classify an unseen instance as either positive or negative. Some assumptions are made in this process, as discussed in Section 2.  

Many methods have been proposed to handle the PU learning task, and as such an exhaustive search of all such methods to find the optimal method for a given learning task would be unfeasible. To address this issue, GA-Auto-PU, based on a Genetic Algorithm (GA) was proposed as the first Automated Machine Learning (Auto-ML) system specific to PU learning \citep{saunders2022ga}. GA-Auto-PU outperformed baseline PU learning methods, including a state-of-the-art approach based on deep forests \citep{zeng2020predicting}. However, the system was very computationally expensive, thus making it potentially impractical for many users. For this reason, in this work we introduce BO-Auto-PU, a Bayesian Optimisation-based Auto-ML system for PU learning, and EBO-Auto-PU, an Evolutionary Bayesian Optimisation-based Auto-ML system for PU learning (a hybrid of GA-Auto-PU and BO-Auto-PU). As shown in Section 6, both systems are much more computationally efficient than GA-Auto-PU and achieve good predictive performance. 

This paper is organised as follows. Section 2 details the background on PU learning and Auto-ML.  Section 3 outlines the Auto-ML framework for PU learning, detailing the search spaces used and the objective function. Section 4 details the three Auto-PU systems, with pseudocodes of the new BO-Auto-PU and EBO-Auto-PU. Section 5 outlines our experimental methodology and the datasets used in the experiments. Section 6 details the results, comparing the three Auto-ML systems amongst themselves and against two baseline PU learning methods. Section 7 concludes this work.

\section{Background}
\subsection{Positive-Unlabelled (PU) Learning}
Positive-Unlabelled (PU) learning is a specialised case of semi-supervised classification that involves training a classifier to distinguish between the positive and negative class, given only positive and unlabelled instances \citep{bekker2020learning}. The aim is to predict an instance’s probability of belonging to the positive class, P(y=1), given information about whether that instance is labelled, P(s=1). It follows that, since each instance is annotated as “labelled positive” or “unlabelled”, 1 or 0, a standard machine learning model will predict those two annotations (positive or unlabelled), rather than whether an instance is positive or negative. Therefore, to accurately predict an instance’s class as positive or negative, alternative approaches must be followed in the place of standard binary classification. 

The most popular PU learning approach is the two-step framework, consisting of a first step to identify reliable negative instances and a second step to learn a classifier that can distinguish the labelled positives and the reliable negatives \citep{bekker2020learning}.

More precisely, in step 1, we generally train a classifier to distinguish between the labelled positive instances and the unlabelled set. Often, this process is completed iteratively, splitting the unlabelled set into multiple sets to handle the class imbalance that is characteristic of many PU learning tasks, where the unlabelled set is usually larger than the positive set \citep{bekker2020learning}. From this process, we learn a classifier that distinguishes between the probability of an instance being labelled, P(s=1), and the probability of an instance being unlabelled, P(s=0). We then assume that those instances with the lowest probability of belonging to the positive class are likely not unlabelled positive instances and are thus treated as the reliable negative set. This assumes smoothness and separability of the data \citep{bekker2020learning}. That is, we assume that instances in the data that are similar have similar probabilities of belonging to the positive class (smoothness) and we assume that there is a natural division between positive and negative classes (separability). 

Many approaches propose a second stage of the first step, akin to a classic semi-supervised learning method, that involves expanding the reliable negative set by training a classifier to distinguish the labelled positives and the initial reliable negative sets, classifying the remaining unlabelled instances, and adding those unlabelled instances with a low P(y=1) to the reliable negative set. 

We refer to the curation of the initial reliable negative set as Phase 1A, and the semi-supervised step as an optional Phase 1B. Note the use of “Phase” rather than “Step”. This allows us to discuss the individual steps involved in each phase without confusion.

The second phase of the framework involves learning a classifier to distinguish the labelled positives and the reliable negative instances (identified in Phase 1). Thus, Phase 2 involves learning a classifier that can predict an instance’s P(y=1) rather than P(s=1).  

Many methods have been proposed in the PU learning literature \citep{bekker2020learning} \citep{jaskie2019positive}. One such early method was the “Spy with Expectation Maximisation” (S-EM) method proposed by \citep{liu2002partially}. S-EM modifies the standard two-step procedure with the introduction of “spy” instances, used to calculate the threshold under which an instance’s P(s=1) must fall in order for the instance to be deemed a reliable negative. S-EM was one of the first two-step methods to be proposed (over two decades ago), but it has proved an effective approach to this day and is frequently used as a baseline method when proposing new PU learning methods \citep{li2010distributional, xia2013instance, ke2018global, zhang2018boosting, schrunner2020generative, liu2021new, he2023novel}. 

A recently proposed state-of-the-art two-step PU learning method is the Deep Forest-PU (DF-PU) method \citep{zeng2020predicting}. DF-PU uses the powerful deep forest classifier \citep{zhou2019deep} in the standard two-step framework to build a PU learning classifier. 

Both S-EM and DF-PU will be used in this work's experiments as baseline methods to evaluate the new Auto-PU systems.

The assumptions most commonly made by PU learning algorithms are 
separability, smoothness, and selected completely at random (SCAR)\citep{bekker2020learning}.

%
%

The assumption of separability states that the positive and negative instances are separable in the feature space. That is, it is assumed that there is a hyperplane that can perfectly separate the positive and negative instances \citep{bekker2020learning}. The assumption of smoothness states that instances which are close to each other in the feature space are likely to belong to the same class \citep{bekker2020learning}. These assumptions are foundational to the two-step approach. 

The SCAR assumption, formalised by \cite{elkan2008learning}, states that the positive instances are labelled irrespective of their features, and thus the labelled set is an independent and identically distributed sample from the positive distribution. That is, for the given data, Pr(s=1) = Pr(s=1 \(\vert\) x), where Pr(s=1) represents the probability of an instance being labelled and x is an instance’s feature vector. Or, put simply, the sample of positive instances in the labelled positive set is representative of the entire set of positive instances, both labelled and unlabelled. Making the SCAR assumption allows us to assume that the instances in the labelled positive set are representative of the instances within the unlabelled set, and thus, if a classifier can accurately predict the labelled positive instances, it should, in theory, be able to predict the unlabelled positive instances also. For this reason, the SCAR assumption is foundational to some PU learning approaches.

The absence of negative instances makes it hard to evaluate PU learning models as predictive accuracy metrics usually rely on knowledge of the true class labels of each instance. However, in PU learning the true class label is known only for a sample of positive instances. The other positive instances, and all negative instances, are unlabelled. Hence, standard accuracy metrics  \citep{japkowicz2011evaluating} cannot be correctly calculated.

Under the SCAR assumption, given that the sample of positive instances in the labelled positive set is representative of the entire set of positive instances, both labelled and unlabelled, we can estimate performance metrics for models tested on genuine PU data; i.e., PU data that has not been engineered from a standard positive-negative (PN) dataset (with positive and negative labels). However, as these metrics are performance estimates, they are not entirely robust. Arguably, a more robust approach is to evaluate a PU learning method on an engineered PU dataset before applying that method to a genuine PU learning task. 

As noted in \citep{saunders2022evaluating}, the approach most frequently used in the literature is to evaluate proposed methods on engineered PU data created from a standard PN dataset by hiding a certain percentage of positive instances in the negative set, thus creating an unlabelled set (i.e., all negatives and the hidden positives will be treated as ‘unlabelled’). This is done for the training set, whilst leaving the test set untouched. That is, the test set will contain positive and negative instances as in the original dataset. Hence, the model is trained on PU data but evaluated on fully labelled data. Therefore, we can accurately calculate standard PN metrics. This is arguably a more robust approach as the SCAR assumption is not required and we can rely on values of performance metrics that are accurately calculated based on the known class labels of the instances in the test set. 

\subsection{Automated Machine Learning (Auto-ML)} 
Automated Machine Learning (Auto-ML) is a rapidly growing field of machine learning that aims to limit human involvement in the machine learning pipeline development process by tailoring such a pipeline to a specific input dataset through the use of an optimisation method \citep{yao2018taking, he2021automl}. This allows users without extensive knowledge of machine learning to operate complex ML pipelines and enables users to find effective pipelines for specific learning tasks without guesswork or arbitrary choices. For a review of Auto-ML approaches, see \citep{zoller2021benchmark, yao2018taking, he2021automl}.

In this work,  we utilise three different types of optimisation approaches for Auto-ML, namely a genetic algorithm (GA), Bayesian optimisation (BO), and a surrogate-assisted evolutionary algorithm, referred to as evolutionary Bayesian optimisation (EBO). 

GAs are optimisation methods that rely on the principle of natural selection to evolve candidate solutions based on a fitness (objective) function \citep{eiben2003introduction}. In Auto-ML, a candidate solution is usually a machine learning (ML) pipeline, and the objective function is usually a predictive accuracy measure  \citep{olson2016evaluation},  
 \citep{de2017recipe}. 
In the context of Auto-ML for PU learning (hereafter named “Auto-PU”), GA-Auto-PU uses a GA for optimisation of PU learning methods \citep{saunders2022ga, Saunders2022EA}. 

BO is an iterative model-based optimisation approach that leverages surrogate models to identify areas of the search space that show promise, either due to their potential predictive performance or due to their placement in an unexplored region of the search space, depending on the acquisition function \citep{frazier2018tutorial}. The acquisition function is a tool to quantity the potential value of evaluating different candidate ML pipelines. The function can be as simple as the value assigned by the surrogate model to the candidate ML pipeline, or a more complex function such as Expected Improvement \citep{mockus1975bayesian} or Probability of Improvement \citep{death2021greed}. Thus, the next candidate solution to evaluate using the objective function is determined by which candidate solution maximises the acquisition function. Generally, the objective function in an Auto-ML setting is expensive as it involves training a ML pipeline. Hence, by selecting a limited number of candidates (often a single candidate solution) to assess with the objective function, computational expense is reduced, making BO a generally much more efficient (faster) optimisation method than a GA. 

One potential drawback of BO is relatively low exploration of the search space and the lack of population diversity (in each BO iteration, typically a single candidate solution is evaluated using the objective function). Both issues are naturally alleviated in GAs through the use of a diverse population and evolutionary operations (crossover and mutation). As such, surrogate-assisted GAs often achieve a good trade-off between the exploration and candidate diversity of a GA and the computational efficiency of BO \citep{jin2011surrogate}. 

In this work, a surrogate model will direct the GA towards promising search space areas whilst maintaining a population to evolve \citep{jin2011surrogate}. Our approach to surrogate-assisted GAs, named Evolutionary Bayesian Optimisation (EBO), is discussed in Section 4. 

\section{Auto-PU: The Proposed Auto-ML Framework for Positive-Unlabelled Learning}
This section outlines the proposed Auto-ML framework for PU learning, specifying the search spaces (Section 3.1) and the objective function used to evaluate candidate solutions (Section 3.2). Note that the search spaces and the objective function are the same for all three types of Auto-PU systems used in this work, which vary regarding their optimisation method: GA, BO or a hybrid EBO. Hence, this common framework of search spaces and objective function is discussed separately in this section, whilst the next three sections discuss the optimisation method-specific details of each type of Auto-PU system.

\subsection{Search Spaces}
In Auto-ML, a search space is the set of all candidate solutions that can be found by the search algorithm, consisting of a pre-defined set of algorithms with their hyperparameters and their respective values. For our Auto-PU systems (Auto-ML systems for PU learning), the search space is defined by the two-step PU learning framework \citep{bekker2020learning}. That is, a candidate solution is a two-step PU learning method, consisting of Phases 1A, 1B, and 2, discussed in Section 2.1. Each phase has a set of hyperparameters and their candidate values, which define the search space of our Auto-PU systems.

Our experiments have used two variations of the search space, namely the base search space and the extended search space, detailed in Subsections 3.1.1 and 3.1.2, respectively.

\subsubsection{The base search space}
The base search space, proposed in our previous work \citep{saunders2022ga}, allows the system to build relatively simple two-step PU learning methods that do not utilise any heuristics for determining the values of the hyperparameters. Specifically, the search space is defined by the following 7 hyperparameters and their corresponding candidate values:
\begin{itemize}
  \setlength\itemsep{0.10em}
    \item \(Iteration\_count\_1A\): \{1, 2, 3, 4, 5, 6, 7, 8, 9, 10 \}
    
    \item \(Threshold\_1A\): \{ 0.05, 0.1, 0.15, 0.2, 0.25, 0.3, 0.35, 0.4, 0.45, 0.5 \}
    
    \item \(Classifier\_1A\): \{ Candidate\_classifiers \}

    \item \(Threshold\_1B\): \{ 0.05, 0.1, 0.15, 0.2, 0.25, 0.3, 0.35, 0.4, 0.45, 0.5 \}
    
    \item \(Classifier\_1B\): \{ Candidate\_classifiers \}
    
    \item \(Flag\_1B\): { True, False }
    
    \item \(Classifier\_2\): \{ Candidate\_classifiers \}
    
\end{itemize}

Where \(Candidate\_classifiers\) represents 18 different candidate classification algorithms: Gaussian naïve Bayes, Bernoulli naïve Bayes, Random forest, Decision tree, Multilayer perceptron, Support vector machine, Stochastic gradient descent classifier, Logistic regression, K-nearest neighbour, Deep forest, AdaBoost, Gradient boosting classifier, Linear discriminant analysis, Extra tree classifier, Extra trees (ensemble) classifier, Bagging classifier,  Gaussian process classifier, and Histogram-based gradient boosting classification tree.  

\begin{figure}
    \centering\includegraphics[width=1.0\textwidth]{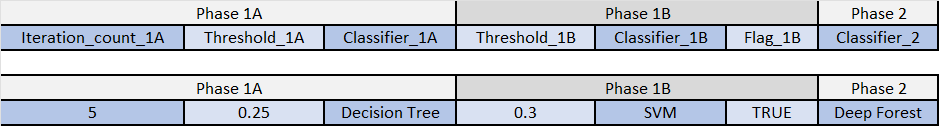}
    \caption{Example of a candidate solution in the base search space.}
    \label{fig1}
\end{figure}

Together, these hyperparameters form Phase 1A, Phase 1B, and Phase 2 of the two-step PU learning framework described in Section 2.1. Figure 1 shows how these hyperparameters form these phases, showing the structure of a candidate solution in the base search space and an example candidate solution (with examples of hyperparameter values). 

Phase 1A consists of the hyperparameters \(Iteration\_count\_1A\), \(Threshold\_1A\), and \(Classifiers\_1A\). \(Iteration\_count\_1A\) specifies the number of subsets into which the unlabelled set is split for learning a classifier to distinguish between the positive and unlabelled sets, and also the number of iterations that a classification algorithm is run in Phase 1A. E.g., if \(Iteration\_count\_1A\) = 5, the unlabelled set is split into 5 subsets, each with 20\% of the unlabelled data, and the classification algorithm is run 5 times, each with a different subset of unlabelled training instances. This helps to handle the class imbalance often found in PU learning datasets. \(Threshold\_1A\) specifies the maximum value of the predicted probability of the positive class for an instance to be considered a reliable negative instance. The \(Classifier\_1A\) is simply the classifier used to predict the reliable negative instances. 

Phase 1B consists of the hyperparameters \(Threshold\_1B\), \(Classifier\_1B\), and \(Flag\_1B\). \(Threshold\_1B\) and \(Classifier\_1B\) are analogous to those used in phase 1A. \(Flag\_1B\) indicates whether or not to skip phase 1B, which is an optional phase in PU learning methods. Given the similarities between Phase 1A and Phase 1B, a natural question arises as to why we exclude an iteration count parameter from Phase 1B. The reason is that the iteration count parameter was introduced in order to handle the class imbalance inherent to PU learning datasets, but this is not generally an issue once an initial reliable negative set has been created as this set is simply a small subset of the unlabelled set. Furthermore, class imbalance is indirectly handled by the \(Threshold\_1A\) parameter, which will probably be adapted to take a smaller value (and thus fewer instances will be added to the reliable negative set) if the reliable negative set becomes large enough to detriment predictive performance.

Phase 2 simply consists of the hyperparameter \(Classifier\_2\). This classifier will be trained to distinguish the positive set and the reliable negative set extracted from the unlabelled set in phase 1A and potentially 1B. 

The base search space has 11,664,000 candidate solutions \citep{saunders2022ga}.

So, in Phase 1A of the example candidate solution shown in Figure 1, the unlabelled set would be split into 5 subsets (defined by the \(Iteration\_count\_1A\) hyperparameter). Each of these subsets in turn, along with the labelled positive set, would be used to train a decision tree classifier (\(Classifier\_1A\)) which would then predict the probability of the unlabelled instances in the current subset belonging to the positive class. Those instances with a predicted probability of less than 0.25 (the \(Threshold\_1A\) hyperparameter) would be added to the reliable negative set and removed from the unlabelled set. Then, as the \(Flag\_1B\) parameter is set to True, a support vector machine (SVM) classifier (\(Classifier\_1B\)) would be built using the labelled positive instances as the positive set and the reliable negative instances identified in Phase 1A as the negative set. It would then be used to classify the remaining unlabelled instances, and those with a predicted probability of belonging to the positive class of less than 0.3 (the \(Threshold\_1B\) hyperparameter) would be added to the reliable negative set. Finally, a deep forest classifier (\(Classifier\_2\)) would be trained on the labelled positive and the reliable negative sets. 

\subsubsection{The extended search space (based on the Spy technique)}
The extended search space, proposed in our work \citep{Saunders2022EA}, introduces three new hyperparameters based on the spy technique for PU learning \citep{liu2002partially}. Spy-based approaches are used to heuristically determine the \(Threshold\_1A\) parameter. A percentage of labelled positive instances (given by \(Spy\_rate\)) are hidden in the unlabelled set as “spy instances”. A classifier (\(Classifier\_1A\)) is built, using the labelled positive instances as the positive set and the unlabelled instances with the spy instances as the negative set. The spy instances are then classified, and \(Threshold\_1A\) is determined such that a percentage of spy instances (given by \(Spy\_tolerance\)) have a predicted probability of belonging to the positive class of less than \(Threshold\_1A\) (e.g., if \(Spy\_tolerance\) = 0.05, 5\% of the spy instances can have a predicted positive-class probability less than \(Threshold\_1A\)). Note that the \(Threshold\_1A\) hyperparameter is thus redundant and its value is not used when building the PU learning model for candidate solutions when \(Spy\_flag\) = True. However,  \(Threshold\_1A\)  is still needed as a component of a candidate solution during the search performed by the Auto-ML system, since some candidate solutions generated along the search will not use the spy technique (depending on the value of the \(Flag\_1B\) hyperparameter). 

The three new hyperparameters introduced into the extended search space are as follows:

\begin{itemize}
  \setlength\itemsep{0.10em}
    \item \(Spy\_flag\): \{ True, False \}
    \item \(Spy\_rate\): \{ 0.05, 0.1, 0.15, 0.2, 0.25, 0.3, 0.35 \}
    \item \(Spy\_tolerance\): \{ 0, 0.01, 0.02, 0.03, 0.04, 0.05, 0.06, 0.07, 0.08, 0.09, 0.1 \}
\end{itemize}

\(Spy\_flag\) is a Boolean value used to indicate whether or not to use a spy-based method in Phase 1A. \(Spy\_rate\) specifies the percentage of positive instances to use as spies. \(Spy\_tolerance\) determines what percentage of spies can remain in the unlabelled set when \(Threshold\_1A\) threshold is calculated.  The inclusion of these three new hyperparameters increases the size of the size space by a factor of 154 (2 × 7 × 11) with respect to the size of the base search space, so the size of the extended search space is 1,796,256,000 candidate solutions.

The motivation for expanding the search space was to attempt to increase predictive performance of the system by utilising spy-based methods, as initially proposed in \citep{liu2002partially}. This approach has been used frequently in the PU learning literature with success \citep{li2010distributional, xia2013instance, ke2018global, zhang2018boosting, schrunner2020generative, liu2021new, he2023novel}. 

Figure 2 shows the representation of a candidate solution in the extended search space, including an example of the value taken by each hyperparameter (specifying an example candidate solution). As the value of \(Spy\_flag\) is set to True, 20\% of the labelled positive instances (determined by \(Spy\_rate\)) are hidden in the unlabelled set in Phase 1A. The RN threshold is determined as the value at which only 1\% of spy instances have a predicted positive-class probability less than the determined value.

\begin{figure}
    \centering
    \includegraphics[width=1.0\textwidth]{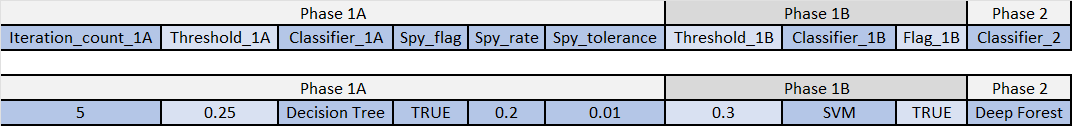}
    \caption{Example of a candidate solution in the extended search space.}
    \label{fig2}
\end{figure}

Spies are utilised in Phase 1A, but not in Phase 1B. This decision was made as preliminary experiments showed no increase in predictive performance when the system allowed spies in Phase 1B. Also, the search space would be greatly expanded if spies were used in Phase 1B, as the three new hyperparameters introduced in this expanded search space would all need to be repeated for Phase 1B, further increasing the size of the search space by another factor of 154. Thus, if spies were used in Phase 1B, the size of the search space would be calculated as 1,796,256,000 × 154, i.e., 276,623,424,000 candidate solutions.

Hence, the use of spies in Phase 1B was not cost-effective in our preliminary experiments, and so we included the spy-based heuristic method in Phase 1A only.

\subsection{Candidate Solution Assessment}

The objective function is used to assess the quality of a given configuration of PU learning hyperparameter settings for a specific PU learning task, i.e., a specific input dataset. This is done by applying the PU method configuration defined by the candidate solution to the training set. This assessment procedure was described in detail in our previous work \citep{saunders2022ga, Saunders2022EA}, but is briefly outlined here for the reader's convenience.

The systems use an internal cross-validation on the training set in order to assess the quality of candidate solutions. The training set is split into 5 folds, and each candidate PU learning algorithm is run 5 times, each time using a different fold as the validation set and the other 4 folds as the learning set. Each time, a candidate two-step PU learning algorithm (specified by the configuration of the candidate solution) is trained on the learning set, and the F-measure of the learned model is assessed on the validation set. That candidate algorithm's quality is the mean F-measure over the 5 validation sets.

The formulas for precision, recall, and F-measure are given in Equations 1, 2 and 3. 

\[
\text{Precision} = \frac{\text{True Positives}}{\text{True Positives} + \text{False Positives}} \quad (1)
\]

\[
\text{Recall} = \frac{\text{True Positives}}{\text{True Positives} + \text{False Negatives}} \quad (2)
\]

\[
\text{F-measure} = \frac{2 \cdot \text{Precision} \cdot \text{Recall}}{\text{Precision} + \text{Recall}} \quad (3)
\]

\noindent Where True Positives are the number of positive instances correctly identified as belonging to the positive class, False Positives are the number of negative instances incorrectly identified as belonging to the positive class, and False Negatives are the number of positive instances incorrectly identified as belonging to the negative class.

\section{Three Auto-ML Systems for Positive-Unlabelled Learning (Auto-PU)}

This section describes the three Auto-PU systems, namely GA-Auto-PU, BO-Auto-PU, and EBO-Auto-PU. Whilst each system utilises a different optimisation method (Genetic Algorithms, Bayesian Optimisation and hybrid Evolutionary/Bayesian Optimisation), all use the same search spaces and objective function, defined in Section 3. Open source implementations of the systems and the baseline methods are available on GitHub
\footnote{https://github.com/jds39/GA-Auto-PU/tree/main}. 

\subsection{GA-Auto-PU}

The GA-Auto-PU system has been described in detail in \citep{saunders2022ga, Saunders2022EA}. Hence, we provide only a brief overview of the system here, for the reader's convenience. 

GA-Auto-PU is a Genetic Algorithm (GA)-based Auto-ML system for PU learning. For an overview of GAs, see Section 2.1. Algorithm 1 outlines the GA procedure. Initially, a Population of \(Pop\_size\) individuals is generated (step 1). The configuration (genome) of the individual is checked against a list of previously assessed configurations, and if it has not already been assessed, the $Fitness$ (predictive accuracy) of $Individual$ is calculated (step 2.a.i.1) as described in Section 3.2. If the configuration has already been assessed, the fitness values of the previous assessment are assigned to the individual (step 2.a.i.2). Once all individuals have been evaluated, the fittest $Individual$ is saved for the next generation (step 2.b). A new population is created from $Population$ after undergoing tournament selection based on fitness (step 2.c), and \(New\_pop\) then undergoes crossover (step 2.d) and mutation (step 2.e) to create new candidate solutions. Finally, $Population$ is set as \(New\_pop\) and \(Fittest\_individual\) (step 2.f). This process of fitness calculation, selection, crossover, mutation, and elitism is repeated \#generations times. The fitness of an individual is the F-measure achieved over the internal 5-fold cross-validation on the training set. 

\begin{algorithm}
	\caption{Outline of the GA procedure} 
	\begin{algorithmic}
        \State 1. $Population = $ Generate population();
        \State 2. \textbf{for} $\#generations$ times \textbf{do}
        \State \hspace{\algorithmicindent}2.a. \textbf{for each} $Individual$ in $Population$ \textbf{do}
        \State \hspace{\algorithmicindent}\hspace{\algorithmicindent}2.a.i. \textbf{if} $Individual$ configuration has not already been assessed \textbf{then}
        \State \hspace{\algorithmicindent}\hspace{\algorithmicindent}\hspace{\algorithmicindent}2.a.i.1. Assess fitness($Individual$, $Training\ set$); 
        \State \hspace{\algorithmicindent}\hspace{\algorithmicindent}2.a.i. \textbf{else}
        \State \hspace{\algorithmicindent}\hspace{\algorithmicindent}\hspace{\algorithmicindent}2.a.i.2. $Individual Fitness$ value is set as the output of the previous assessment;
        \State \hspace{\algorithmicindent}2.b. $Fittest\_individual = $ Get fittest individual($Population$);
        \State \hspace{\algorithmicindent}2.c. $New\_pop = $ 100 individuals selected from $Population$ by tournament selection;
        \State \hspace{\algorithmicindent}2.d. Individuals in $New\_pop$ undergo crossover;
        \State \hspace{\algorithmicindent}2.e. Individuals in $New\_pop$ undergo mutation;
        \State \hspace{\algorithmicindent}2.f. $Population = New\_pop \cup \{Fittest\_individual\}$;
        \State 3. \textbf{Return} $Individual$ in $Population$ with the highest fitness;
	\end{algorithmic} 
\end{algorithm}

The system utilises standard uniform crossover and mutation (replacing a gene’s value by a randomly sampled candidate value) as search operators. 

\subsection{BO-Auto-PU}

In this section we introduce BO-Auto-PU, a new Bayesian Optimisation (BO)-based Auto-ML system for PU learning. GA-Auto-PU, described in Section 4.1, was the first Auto-ML system specific to PU learning, and showed statistically significant improvements in predictive accuracy over PU learning baselines \citep{saunders2022ga, Saunders2022EA}. However, GA-Auto-PU is computationally expensive, averaging about 225 minutes to run 5-fold cross-validation per dataset \citep{saunders2022ga, Saunders2022EA}.

BO is generally a much more computationally efficient procedure than a standard GA, given that it assesses most of the generated individuals (candidate solutions) with a fast executed surrogate model as opposed to the slowly executed objective function. As such, BO-Auto-PU has been developed in an attempt to reduce the computational expense of GA-Auto-PU without sacrificing predictive accuracy. Algorithm 2 outlines the BO procedure.

Initially, \(\#Configs\) PU learning algorithm configurations are randomly generated (step 1) and evaluated, with their F-measures saved as Scores (step 2). A random forest regressor, \(Surr\_model\) (surrogate model), is then trained using $Configs$ as features and Scores as the target variable (step 3). In each iteration, a new set of \(\#Configs\) configurations, \(New\_configs\), are randomly generated (step 4.a) and a surrogate score for each is calculated by \(Surr\_model\) and saved as \(\hat{Y}\) (step 4.b). The best configuration, \(Best\_config\), with the highest score according to \(\hat{Y}\) is evaluated using the objective function (steps 4.c,d), and added to $Configs$, with the objective Score (F-measure) added to $Scores$ (step 4.e). \(Surr\_model\) is then retrained with \(Best\_config\) added to $Config$ (step 4.f). Steps 4.a-f are repeated \(It\_count\) times. Finally, the configuration with the best objective score is returned. 

$Configs$ are processed as follows for training \(Surr\_model\): for the base search space (without the Spy method), the numeric components of each configuration (\(Threshold\_1A\), \(Iteration\_Count\_1A\), \(Threshold\_1B\)) are treated as numeric features, the Boolean component (\(Flag\_1B\)) is treated as a binary feature, and the nominal components (\(Classifier\_1A\), \(Classifier\_1B\), \(Classifier\_2\)) are one-hot encoded, with a binary value for each potential value of the component, indicating whether or not that value is used. The resulting instance for the regression algorithm (which will learn \(Surr\_model\)) has 58 features. For the extended search space (with the Spy method), all the previously mentioned components are treated as they are in the base search space. However, we also have the Spy method’s components, with \(Spy\_rate\) and \(Spy\_tolerance\) treated as numeric features, and the Boolean component “Spies” treated as a binary feature –- a flag, indicating whether or not the Spy method is used. This results in an instance with 61 features for the regression algorithm.  

\begin{algorithm}
    \caption{Outline of the BO procedure for Positive-Unlabelled learning} 
    \begin{algorithmic}
        \State 1. $Configs$ = randomly generate \(\#Configs\) PU learning configurations;
        \State 2. $Scores$ = run objective function for all configurations in $Configs$;
        \State 3. Fit \(Surr\_model\) with $Configs$ as features and $Scores$ as the target;
        \State 4. \textbf{for} $\#It\_count$ times \textbf{do}
            \State \hspace{\algorithmicindent}4.a. \(New\_configs\) = randomly generate \(\#Configs\) configurations;
            \State \hspace{\algorithmicindent}4.b. \(\hat{Y}\) = calculate a surrogate score for each new config with \(Surr\_model\);
            \State \hspace{\algorithmicindent}4.c. \(Best\_config\) = config with the highest score according to \(\hat{Y}\);
            \State \hspace{\algorithmicindent}4.d. $Score$ = run the objective function for \(Best\_config\);
            \State \hspace{\algorithmicindent}4.e. Add \(Best\_config\) to $Configs$, add $Score$ to $Scores$;
            \State \hspace{\algorithmicindent}4.f. Retrain \(Surr\_model\) on $Configs$ and $Scores$;
        \State 5. \textbf{Return} the Best configuration according to the objective score;
    \end{algorithmic} 
\end{algorithm}

Table \ref{tab2} shows the hyperparameter settings of the BO underlying BO-Auto-PU. The \(It\_count\) parameter determines the number of iterations to perform the optimisation. \(\#Configs\) determines the number of individuals in the population. \(Surr\_model\) is the surrogate model used to calculate the surrogate score. The acquisition function is the method for selecting which candidate solution to assess with the objective function. For this, we simply use the surrogate score calculated by \(Surr\_model\). The \(It\_count\) and \(\#Configs\) parameters were set to match the \(\#generations\) and \(Pop\_size\) parameters for GA-Auto-PU. In Bayesian optimisation, it is common to use an acquisition function that attempts to trade-off exploration and exploitation. However, in our case, preliminary experiments showed worse performance utilising these acquisition functions than simply utilising the surrogate score. As such, we have used the surrogate score only.

\begin{table}[]
\caption{Hyperparameters of the BO-Auto-PU system, with their default values.}\label{tab2}
\centering
\small
\begin{tabular}{|l|l|}
\hline
Hyperparameter       & Value                        \\\hline
\(It\_count\)             & 50                           \\\hline
\(\#Configs\)             & 101                          \\\hline
\(Surr\_model\)           & Random   Forest Regressor    \\\hline
\(Acquisition function\) & \(Surr\_model\)   predicted value \\ \hline
\end{tabular}
\end{table}

\subsection{EBO-Auto-PU}

As mentioned earlier, GA-Auto-PU's performance came at a large computational cost, averaging about 225 minutes for a 5-fold cross-validation per dataset. In an effort to reduce the computational cost of GA-Auto-PU, we introduced BO-Auto-PU, detailed in the previous section. BO-Auto-PU proved much more computationally efficient than GA-Auto-PU, with BO-Auto-PU averaging less than 10 minutes to run a 5-fold cross-validation per dataset. 

The improvement in computational efficiency, however, was gained at a small cost to predictive performance, as shown later in this work. Of the four Auto-ML systems tested (GA and BO, each with two versions (two search spaces)), as shown in the results reported later, the GA with base space was arguably the best system in terms of predictive accuracy, whilst BO with the extended search space was undoubtedly the worst system. It seems that the BO-based system was unable to cope with the expanded search space. 

We have hypothesised that the disparity in predictive accuracy was due to the differing levels of population diversity between the two systems. That is, GA-Auto-PU has large population diversity as it assesses many configurations in each iteration according to the slow objective function, and it creates diversity through crossover and mutation applied to configurations selected based on their fitness (predictive accuracy). Whereas BO-Auto-PU assesses only a single configuration at each iteration according to the slow objective function, namely a configuration that is selected according to the predictive accuracy value predicted by the random forest regressor. At each iteration, a population is randomly generated to be evaluated by the faster surrogate model, rather than evolved to be evaluated by the slower objective (fitness) function as in the case of GA-Auto-PU. So, whilst diversity may occur in the BO search, there is no guarantee that it will benefit the system as the diverse configurations may not be selected for assessment by the objective function. 

EBO-Auto-PU was developed in an attempt to bridge this gap between the GA- and the BO-based systems, introducing diversity into BO by evolving a population rather than random population generation, whilst utilising a surrogate model to reduce computational expense and prioritise candidate solutions for assessment according to the objective function. The EBO procedure is akin to a surrogate-assisted genetic algorithm (SAGA), with the difference being the number of candidate solutions selected for evaluation (1 + $k$, as opposed to the normal 1 in a SAGA) and the manner of selection of the \(k\_pop\) sample, selected with tournament selection and with genetic operators applied to.

Algorithm 3 outlines the procedure that the new EBO component of EBO-Auto-PU follows to evolve a PU learning algorithm. \(\#Configs\) PU learning configurations are randomly generated (step 1) and evaluated, with their F-measures saved as Scores (step 2). A random forest regressor, \(Surr\_model\), is then trained, using $Configs$ as features, and $Scores$ as the target variable (step 3). The configurations in $Configs$ are copied to a temporary store $Temp\_Configs$ to undergo crossover and mutation to produce an evolved population of configurations (step 4.a). The surrogate scores of each just-produced configuration in $Configs$ is calculated by \(Surr\_model\) and saved as \(\hat{Y}\)(step 4.b), with the configuration with the highest surrogate score saved as \(Best\_config\) (step 4.c). Tournament selection is utilised to select $k$ candidate solutions according to their objective score (F-measure), which are then added to a population \(k\_pop\) (step 4.d-e). \(k\_pop\) then undergoes crossover and mutation to produce a newly evolved population (step 4.f). \(Best\_config\) and the configurations from \(k\_pop\) are assessed according to the objective function to calculate their objective scores, before the configurations are added to $Configs$ and used to update the \(Surr\_model\) (steps 4.g-i). The objective function cited in step 4.g is defined in Section 3.2. This process (step 4 in Algorithm 3) is repeated \(It\_count\) times. Finally, the best configuration, according to the objective score, is returned. 

Note that the computation of the objective score is much more expensive than the computation of the surrogate score, and hence, at each iteration, just k + 1 configurations (\(Best\_config\) and the $k$ configurations in \(k\_pop\)) have their objective score computed.

\begin{table}[]
\caption{Hyperparameters of the EBO-Auto-PU system, with their default values.}\label{tab3}
\centering
\small
\begin{tabular}{|l|l|}
\hline
Hyperparameter                    & Value                   \\ \hline
\(\#Configs\)                          & 101                     \\ \hline
\(It\_count\)                          & 50                      \\ \hline
\(Surr\_model\)                        & Random Forest Regressor \\ \hline
Crossover probability             & 0.9                     \\ \hline
Component crossover probability    & 0.5                     \\ \hline
Mutation probability              & 0.1                     \\ \hline
Tournament size                   & 2                       \\ \hline
\(k\)                                 & 10                      \\ \hline
\end{tabular}
\end{table}

$Configs$ are processed as follows for training \(Surr\_model\): for the base search space, the numeric components of each configuration (\(Threshold\_1A\), \(Iteration\_Count\_1A\), \(Threshold\_1B\)) are treated as numeric features, the Boolean component (\(Flag\_1B\)) is treated as a binary feature, and the nominal components (\(Classifier\_1A\), \(Classifier\_1B\), \(Classifier\_2\)) are one-hot encoded, with a binary value for each potential value of the component, indicating whether or not that value is used by the PU learning method. The resulting instances used as input by the regression algorithm consist of 58 features. For the extended search space, all the previously mentioned components are treated as they are in the base search space. However, we also have the additional spy components, with \(Spy\_rate\) and \(Spy\_tolerance\) treated as numeric features, and the Boolean component “Spies” treated as a binary feature. This results in instances consisting of 61 features.  

EBO-Auto-PU's hyperparameters and their candidate values are shown in Table \ref{tab3}.  Most of these hyperparameters are inherited from GA-Auto-PU and BO-Auto-PU, as follows.

The hyperparameters \(\#Configs\), \(It\_count\) and \(Surr\_model\) in Table \ref{tab3} are the same as the corresponding hyperparameters in Table \ref{tab2} for the BO procedure, and these hyperparameters take the same settings in both tables (for both EBO and BO), in order to make the comparison between EBO-Auto-PU and BO-Auto-PU as fair as possible. The hyperparameters “Crossover probability”, “Component crossover probability”, “Mutation probability”, and “Tournament size” in Table \ref{tab3} are also the same as the hyperparameters “\(Cross\_prob\)”, “\(Gene\_cross\_prob\)”, “\(Mutation\_prob\)”, and “\(Tournament\_size\)” for the GA-Auto-PU system, respectively; and again, these hyperparameters take the same settings for a fair comparison between EBO-Auto-PU and GA-Auto-PU. The only parameter unique to EBO-Auto-PU is the $k$ parameter, used to determine the number of candidate solutions to be selected with tournament selection to assess according to their objective score.

\begin{algorithm}
    \caption{Outline of the EBO procedure for Positive-Unlabelled learning} 
    \begin{algorithmic}
        \State 1. $Configs$ = randomly generate \(\#Configs\) PU learning configurations;
        \State 2. $Scores$ = run objective function for all configurations in $Configs$;
        \State 3. Fit \(Surr\_model\) with $Configs$ as features and $Scores$ as the target;
        \State 4. \textbf{for} $\#It\_count$ times \textbf{do}
        \State \hspace{\algorithmicindent}4.a. $Temp\_Configs$ = $Configs$ after undergoing crossover and mutation;
        \State \hspace{\algorithmicindent}4.b. \(\hat{Y}\) = calculate a surrogate score for each new config in $Temp_Configs$ with \(Surr\_model\);
        \State \hspace{\algorithmicindent}4.c. \(Best\_config\) = config with the highest score according to \(\hat{Y}\);
        \State \hspace{\algorithmicindent}4.d. \(k\_pop\) = \{ \};
        \State \hspace{\algorithmicindent}4.e. \textbf{for} $k$ times \textbf{do}
        \State \hspace{\algorithmicindent}\hspace{\algorithmicindent}4.e.i. \(k\_cand\_sol\) = select from $Temp_Configs$ using tournament selection based on surrogate scores;
        \State \hspace{\algorithmicindent}\hspace{\algorithmicindent}4.e.ii. \(k\_pop\) = \(k\_pop \cup k\_cand\_sol\);
        \State \hspace{\algorithmicindent}4.f. \(k\_pop\) = configurations in \(k\_pop\) undergo crossover and mutation;
        \State \hspace{\algorithmicindent}4.g. Assess \(Best\_config\) and each configuration from \(k\_pop\) with objective function to obtain objective scores;
        \State \hspace{\algorithmicindent}4.h. $Configs$ = \(Configs \cup Best\_config \cup k\_pop\);
        \State \hspace{\algorithmicindent}4.i. Retrain \(Surr\_model\) on $Configs$;
        \State 5. \textbf{Return} Best configuration according to the objective score;
    \end{algorithmic} 
\end{algorithm}

\subsection{Computational Efficiency}

In the GA-based, BO-based, and EBO-based Auto-ML systems for PU learning, the running time is by far dominated by the time to evaluate the candidate solutions along the iterations of the search, i.e., the time to learn a PU model and evaluate its F-measure on the training set, for each candidate PU learning method. GA, BO, and the EBO-based methods perform the same number of iterations (50) in our experiments. However, in each generation (iteration) of GA-Auto-PU the GA must learn $n$ PU models, where $n$ is the number of individuals (candidate solutions) in the population; each iteration of BO-Auto-PU needs to learn a single PU model; whilst each iteration of EBO-Auto-PU needs to learn $k$+1 models, where $k$ is the number of candidate solutions selected via tournament selection to assess according to the objective function. Learning a PU model can be very computationally expensive, depending not only on the size of the dataset but also on the time complexity of the 3 classification algorithms chosen for Phases 1A, 1B, and 2 of the 2-step method, and the number of iterations the classifier is applied in Phase 1A.

All three Auto-ML systems also must perform other steps for generating candidate solutions to be evaluated, but these take in general much less time than the time to evaluate candidate solutions using the objective function (F-measure) as described above. More precisely, at each iteration the GA and the EBO must perform tournament selection, crossover and mutation, but these are all simple operations, which are much faster than computing the fitness function (learning one PU model for each individual).

Unlike the GA, at each iteration BO and EBO learn a surrogate model, but again, the time for this is much shorter than the time taken to learn a PU model in each iteration of BO. This is because the surrogate model is learned by a relatively fast random forest algorithm using a small dataset of PU algorithm configurations, whilst learning a PU model involves running multiple classifiers (one of them for several iterations in phase I-A), each classifier can be much slower than a random forest. In addition, each classifier is learned using the training data of the current dataset, which is typically much larger in number of instances than the small dataset of PU method configurations. As EBO-Auto-PU assesses (using the expensive objective function) fewer candidate solutions than GA-Auto-PU, but more than BO-Auto-PU, the EBO-Auto-PU system sits between the GA-Auto-PU and the BO-Auto-PU systems in regard to computational runtime. More precise results about the differences of computational time among these three systems will be reported in Section 6.
\section{Experimental Methodology}

\subsection{Datasets Used in the Experiments}
Experiments were conducted on 20 publicly available biomedical datasets, including 13 classical benchmark classification datasets from the well-known UCI dataset repository \citep{asuncion2007uci}, and 7 datasets from \citep{marcus2010open, pereira2016somatic, fleming1991counting, islam2020likelihood, chicco2020machine, hlavnicka2017automated, emon2020performance}. 
These datasets all involve real-world learning scenarios of disease or health-risk prediction. The main characteristics of these datasets (numbers of instances and features, and percentage of positive instances) are shown in Table \ref{tab1}. 

These datasets are originally binary classification datasets and so need to be adapted for PU learning. To do so, in each training set, we have hidden \(\delta\)\% of the positive instances in the negative set (where \(\delta\) is a user-specified parameter), thus creating an unlabelled set. This process of engineering a PU dataset from a binary dataset is common throughout the PU learning literature \citep{liu2002partially, he2010naive, basile2017density, saunders2022evaluating}. \(\delta\) takes the values 20\%, 40\%, and 60\% throughout this work, meaning that each dataset is engineered into three datasets, thus creating 60 datasets total. Due to the nature of PU learning, it is often hard to know the distribution of positive instances, and what proportion of them remain unlabelled. Hence, to further analyse the performance of a PU learning algorithm, it is important, when feasible, to assess the performance of its learned model on multiple distributions of unlabelled instances. That is, testing on different versions of the same dataset with differing percentages of the positive instances hidden in the unlabelled set in the training set, when doing experiments with engineered PU datasets.

\begin{table}[htbp]
\caption{Main characteristics of the biomedical datasets used in the experiments.}\label{tab1}
\centering
\small
\begin{tabular}{|l|c|c|c|}
\hline
\textbf{Dataset} & \multicolumn{1}{l|}{\textbf{No. inst.}} & \multicolumn{1}{l|}{\textbf{No. feat.}} & \multicolumn{1}{l|}{\textbf{Posit.-class \%}} \\ \hline
\begin{tabular}[c]{@{}l@{}}Alzheimer's  \citep{marcus2010open} \end{tabular}                   & 354  & 9  & 10.73\% \\ \hline
\begin{tabular}[c]{@{}l@{}}Autism   \citep{asuncion2007uci} 
\end{tabular}                      & 288  & 15 & 48.26\% \\ \hline
\begin{tabular}[c]{@{}l@{}}Breast cancer Coimbra 
 \citep{asuncion2007uci}
\end{tabular}        & 116  & 9  & 55.17\% \\ \hline
\begin{tabular}[c]{@{}l@{}}Breast cancer Wisconsin 
 \citep{asuncion2007uci}
\end{tabular}      & 569  & 30 & 37.26\% \\ \hline
\begin{tabular}[c]{@{}l@{}}Breast cancer mutations  \citep{pereira2016somatic}\end{tabular}   & 1416 & 53 & 32.42\% \\ \hline
\begin{tabular}[c]{@{}l@{}}Cervical cancer 
 \citep{asuncion2007uci}
\end{tabular}              & 668  & 30 & 2.54\%  \\ \hline
\begin{tabular}[c]{@{}l@{}}Cirrhosis  \citep{fleming1991counting}\end{tabular}                & 277  & 17 & 25.72\% \\ \hline
\begin{tabular}[c]{@{}l@{}}Dermatology 
 \citep{asuncion2007uci}
\end{tabular}                 & 359  & 34 & 13.41\% \\ \hline
\begin{tabular}[c]{@{}l@{}}Pima Indians Diabetes 
 \citep{asuncion2007uci}
\end{tabular}        & 769  & 8  & 34.90\% \\ \hline
\begin{tabular}[c]{@{}l@{}}Early Stage Diabetes  \citep{islam2020likelihood}\end{tabular}     & 521  & 17 & 61.54\% \\ \hline
\begin{tabular}[c]{@{}l@{}}Heart Disease  \citep{asuncion2007uci}  \end{tabular}              & 304  & 13 & 54.46\% \\ \hline
\begin{tabular}[c]{@{}l@{}}Heart Failure  \citep{chicco2020machine} \end{tabular}             & 300  & 12 & 32.11\% \\ \hline
\begin{tabular}[c]{@{}l@{}}Hepatitis C  \citep{asuncion2007uci} \end{tabular}                 & 590  & 13 & 9.51\%  \\ \hline
\begin{tabular}[c]{@{}l@{}}Kidney Disease  \citep{asuncion2007uci} \end{tabular}              & 159  & 24 & 27.22\% \\ \hline
\begin{tabular}[c]{@{}l@{}}Liver Disease  \citep{asuncion2007uci} \end{tabular}               & 580  & 11 & 71.50\% \\ \hline
\begin{tabular}[c]{@{}l@{}}Maternal Risk  \citep{asuncion2007uci} \end{tabular}               & 1014 & 6  & 26.82\% \\ \hline
\begin{tabular}[c]{@{}l@{}}Parkinsons  \citep{asuncion2007uci} \end{tabular}                  & 196  & 22 & 75.38\% \\ \hline
\begin{tabular}[c]{@{}l@{}}Parkinsons Biomarkers  \citep{hlavnicka2017automated}\end{tabular} & 131  & 29 & 23.08\% \\ \hline
\begin{tabular}[c]{@{}l@{}}Spine  \citep{asuncion2007uci}  \end{tabular}                      & 311  & 6  & 48.39\% \\ \hline
\begin{tabular}[c]{@{}l@{}}Stroke  \citep{emon2020performance} \end{tabular}                  & 3427 & 15 & 5.25\%  \\ \hline
\end{tabular}
\end{table}

Biomedical datasets are good candidates for PU learning given the inherent uncertainty involved in labelling biomedical data. For example, when learning whether or not a person has a given disease (the scenario for many of ourdatasets), a classifier may learn to distinguish between a positive set (patients diagnosed with that disease) and a negative set 
(patients not diagnosed with that disease). From this data, we wish to identify whether an unseen patient \textit{has} a specific disease. However, consider the wording of this scenario. We are looking to identify whether a patient has a disease, by learning from data consisting of patients who have or have not been diagnosed with a disease. In other words, we are looking to identify true positives by learning only from labelled positives. The apparently negative set, in this scenario, can be more precisely considered an unlabelled set, given that “not diagnosed” does not mean that a patient does not have a disease. It might simply be that this patient has not undergone any tests to determine whether the disease is present. Or, this patient may have undergone some tests, but the disease may be undetectable with the given test. For examples of studies detailing the reliability of specific diagnostic tests see \citep{shinkins2017diagnostic, beach2018importance, debruyn2001systematic, palmedo2006integrated, nerad2016diagnostic, zhang2017meta}. Furthermore, biomedical tests are expensive, and thus the presence of unlabelled data may simply be a practicality to minimise the cost of data curation. Thus, we have used biomedical datasets in our experiments as they are appropriate for PU learning and have been referred to as “one of the most significant usage areas in PU learning” \citep{jaskie2022positive}.

\subsection{Nested Cross-validation}
Throughout this work, the experiments use a nested cross-validation procedure, with an external cross-validation used to measure predictive performance and an internal cross-validation used to evaluate candidate solutions during a run of an Auto-PU system. 

For the external cross-validation, the experiments use the well-known stratified 5-fold cross-validation procedure. This involves randomly splitting the data into 5 folds, and then using one fold at a time as the test set and the other four folds as the training set. The predictive performance of a method is its average performance over the 5 test sets in this external cross-validation. The cross-validation is stratified in the sense that in each of the 5 folds the distribution of class labels is approximately the same as the distribution in the full dataset. We chose 5 folds, rather than the more popular 10 folds, as the number of positive instances in some of our datasets are small. Thus, in some datasets 10 folds would split the positive set into folds that are so small as to be practically unsuitable. 

For each iteration of the external cross-validation (i.e. for each pair of training and test sets), we run a stratified internal cross-validation, where the training set is randomly split into 5 folds and the Auto-PU system is run 5 times, using one fold at a time as the validation set (to measure performance) and the other 4 folds as the learning set (to learn a classifier using a candidate PU learning algorithm, out of the algorithms in the Auto-PU system's search space). The performance of a candidate PU learning algorithm is determined by the average F-measure value achieved over the 5 validation sets. When these 5 runs of the Auto-PU system are over, the best PU learning algorithm found by the system is thus the one with the highest average F-measure over the 5 validation sets of the current training set. Then, a PU learning classifier is built from the full training set with the configuration defined by that best PU learning algorithm. The classifier is then used to predict the class of all instances in the test set of the current iteration of the external cross-validation. This process is repeated for the 5 pairs of training and test sets in the external 5-fold cross-validation. 

When comparing a version of the Auto-PU system against another version or a PU learning method, all systems/methods tested use the same nested cross-validation procedure, with the same folds, to ensure a fair comparison. 

\subsection{Statistical Significance Analysis}
For each performance measure (F-measure, recall, and precision), we compare the performance of an Auto-ML system against the performance of other Auto-ML systems or baseline PU learning methods using the non-parametric Wilcoxon Signed-Rank test \citep{wilcoxon1963critical}. Since this involved testing multiple null hypotheses, we use the well-known Holm correction \citep{demvsar2006statistical} for multiple hypothesis testing. This procedure involves comparing the best method against each of the other methods, ranking the p-values from the smallest to largest (i.e., from most to least significant), and adjusting the significance level \(\alpha\) according to the p-values’ ranking. We set \(\alpha\) = 0.05 as usual before adjusting it according to the position of the p-value in the ranked list. \(p_1\) (the smallest p-value) is deemed significant if it is less than \(\alpha / n\), where \(n\) is the number of hypotheses tested, which is the number of methods tested minus 1. For example, \(n\) = 2 for 3 methods, since the best method is compared against each of the other two methods, and so 2 hypotheses are tested. If this condition is not satisfied, the procedure stops and all p values are deemed non-significant. If \(p_1\) is deemed significant, \(p_2\) is deemed significant if it is less than \(\alpha/(n-1)\), etc.

\subsection{Hyperparameter optimisation for the baseline PU learning methods}

Since the Auto-PU systems optimise the configuration of a PU learning algorithm, we also optimised the hyperparameters of the baseline PU learning methods (S-EM, DF-PU) with a grid search (via an internal cross-validation on the training set), for a fair comparison. For DF-PU, we optimised the percentage of unlabelled instances selected as reliable negatives (reliable negative rate) and the number of iterations, with the following  candidate values:

\begin{itemize}
  \setlength\itemsep{0.10em}    \item Reliable negative rate: \{ 0.01, 0.03, 0.05, 0.07, 0.09, 0.11, 0.13, 0.15, 0.17, 0.2 \}
    \item Iteration count: \{ 1, 2, 3, 4, 5, 6, 7, 8, 9, 10 \}
\end{itemize}

For S-EM, the hyperparamters optimised were the same as those optimised for the extended search space of the Auto-PU systems; namely the spy rate and the spy tolerance. The candidate values of these hyperparameters were as follows.

\begin{itemize}
  \setlength\itemsep{0.10em}    \item Spy rate: \{ 0.05, 0.1, 0.15, 0.2, 0.25, 0.3, 0.35 \}
    \item Spy tolerance: \{ 0, 0.01, 0.02, 0.03, 0.04, 0.05, 0.06, 0.07, 0.08, 0.09, 0.1 \}
\end{itemize}

\section{Results and Discussion}

This section consists two parts. Section 6.1 reports the results for the base search space, and Section 6.2 reports the results for the extended search space. Each of these sections is further divided into two parts, first comparing the Auto-PU systems among themselves and then comparing the Auto-PU systems against  two baseline PU learning methods, namely  S-EM \citep{liu2002partially} and DF-PU \citep{zeng2020predicting} (briefly reviewed in Section 2.1). This section reports the F-measure results per dataset, as this is the most popular metric in PU learning \citep{saunders2022evaluating}.
The precision and recall results per dataset are not reported to save space; but this section reports statistical significance results for all three measures (F-measure, precision, recall), for all datasets as a whole.


Hereafter we will use the short names GA-1, BO-1 and EBO-1 to refer to the versions of GA-Auto-PU, BO-Auto-PU and EBO-Auto-PU with the $base$ search space; and use the short names GA-2, BO-2 and EBO-2 to refer to the versions of GA-Auto-PU, BO-Auto-PU and EBO-Auto-PU with the $extended$ search space. We will also use the term Auto-PU to refer to the Auto-ML systems for PU learning.

\subsection{Results for the Three Auto-PU systems with the base search space}

Recall that the base search space allows for simple PU learning algorithms to be developed  without spy-based methods (which are included only in the extended search space). 

\subsubsection{Comparing the Three Auto-PU Systems}

Table \ref{tab4} reports the F-measure values achieved by the three Auto-PU systems with the base search space on each of the 20 datasets, for \(\delta\) = 20\%, 40\%, 60\%. 
Recall that \(\delta\) denotes the percentage of positive instances hidden in the unlabelled set.

\begin{table}[htbp]
\caption{F-measure results of the Auto-PU systems utilising the base search space.}\label{tab4}
\centering
\small
\begin{tabular}{|l|ccc|ccc|ccc|}
\hline
\multirow{2}{*}{Dataset} &
  \multicolumn{3}{c|}{\(\delta\) = 20\%} &
  \multicolumn{3}{c|}{\(\delta\) = 40\%} &
  \multicolumn{3}{c|}{\(\delta\) = 60\%} \\ \cline{2-10} 
 &
  \multicolumn{1}{l|}{EBO-1} &
  \multicolumn{1}{l|}{GA-1} &
  BO-1 &
  \multicolumn{1}{l|}{EBO-1} &
  \multicolumn{1}{l|}{GA-1} &
  BO-1 &
  \multicolumn{1}{l|}{EBO-1} &
  \multicolumn{1}{l|}{GA-1} &
  BO-1 \\ \hline
Alzheimer’s &
  \multicolumn{1}{l|}{0.629} &
  \multicolumn{1}{l|}{0.529} &
  0.615 &
  \multicolumn{1}{l|}{0.587} &
  \multicolumn{1}{l|}{0.551} &
  0.600 &
  \multicolumn{1}{l|}{0.540} &
  \multicolumn{1}{l|}{0.456} &
  0.436 \\ \hline
Autism &
  \multicolumn{1}{l|}{0.986} &
  \multicolumn{1}{l|}{0.960} &
  0.967 &
  \multicolumn{1}{l|}{0.926} &
  \multicolumn{1}{l|}{0.927} &
  0.956 &
  \multicolumn{1}{l|}{0.927} &
  \multicolumn{1}{l|}{0.910} &
  0.863 \\ \hline
Breast cancer Coi. &
  \multicolumn{1}{l|}{0.966} &
  \multicolumn{1}{l|}{0.705} &
  0.694 &
  \multicolumn{1}{l|}{0.952} &
  \multicolumn{1}{l|}{0.687} &
  0.701 &
  \multicolumn{1}{l|}{0.615} &
  \multicolumn{1}{l|}{0.510} &
  0.586 \\ \hline
Breast cancer Wis. &
  \multicolumn{1}{l|}{0.893} &
  \multicolumn{1}{l|}{0.954} &
  0.949 &
  \multicolumn{1}{l|}{0.872} &
  \multicolumn{1}{l|}{0.932} &
  0.969 &
  \multicolumn{1}{l|}{0.927} &
  \multicolumn{1}{l|}{0.906} &
  0.895 \\ \hline
Breast cancer mut. &
  \multicolumn{1}{l|}{0.672} &
  \multicolumn{1}{l|}{0.893} &
  0.893 &
  \multicolumn{1}{l|}{0.667} &
  \multicolumn{1}{l|}{0.868} &
  0.873 &
  \multicolumn{1}{l|}{0.862} &
  \multicolumn{1}{l|}{0.854} &
  0.841 \\ \hline
Cervical cancer &
  \multicolumn{1}{l|}{0.839} &
  \multicolumn{1}{l|}{0.828} &
  0.839 &
  \multicolumn{1}{l|}{0.904} &
  \multicolumn{1}{l|}{0.903} &
  0.903 &
  \multicolumn{1}{l|}{0.667} &
  \multicolumn{1}{l|}{0.714} &
  0.645 \\ \hline
Cirrhosis &
  \multicolumn{1}{l|}{0.532} &
  \multicolumn{1}{l|}{0.573} &
  0.545 &
  \multicolumn{1}{l|}{0.453} &
  \multicolumn{1}{l|}{0.464} &
  0.529 &
  \multicolumn{1}{l|}{0.507} &
  \multicolumn{1}{l|}{0.443} &
  0.489 \\ \hline
Dermatology &
  \multicolumn{1}{l|}{0.899} &
  \multicolumn{1}{l|}{0.860} &
  0.872 &
  \multicolumn{1}{l|}{0.813} &
  \multicolumn{1}{l|}{0.780} &
  0.905 &
  \multicolumn{1}{l|}{0.716} &
  \multicolumn{1}{l|}{0.828} &
  0.725 \\ \hline
PI Diabetes &
  \multicolumn{1}{l|}{0.654} &
  \multicolumn{1}{l|}{0.677} &
  0.647 &
  \multicolumn{1}{l|}{0.661} &
  \multicolumn{1}{l|}{0.649} &
  0.645 &
  \multicolumn{1}{l|}{0.634} &
  \multicolumn{1}{l|}{0.606} &
  0.594 \\ \hline
ES Diabetes &
  \multicolumn{1}{l|}{0.973} &
  \multicolumn{1}{l|}{0.958} &
  0.983 &
  \multicolumn{1}{l|}{0.913} &
  \multicolumn{1}{l|}{0.895} &
  0.877 &
  \multicolumn{1}{l|}{0.909} &
  \multicolumn{1}{l|}{0.930} &
  0.902 \\ \hline
Heart Disease &
  \multicolumn{1}{l|}{0.833} &
  \multicolumn{1}{l|}{0.843} &
  0.844 &
  \multicolumn{1}{l|}{0.800} &
  \multicolumn{1}{l|}{0.801} &
  0.830 &
  \multicolumn{1}{l|}{0.774} &
  \multicolumn{1}{l|}{0.785} &
  0.777 \\ \hline
Heart Failure &
  \multicolumn{1}{l|}{0.732} &
  \multicolumn{1}{l|}{0.770} &
  0.753 &
  \multicolumn{1}{l|}{0.666} &
  \multicolumn{1}{l|}{0.652} &
  0.605 &
  \multicolumn{1}{l|}{0.640} &
  \multicolumn{1}{l|}{0.674} &
  0.704 \\ \hline
Hepatitis C &
  \multicolumn{1}{l|}{0.925} &
  \multicolumn{1}{l|}{0.953} &
  0.907 &
  \multicolumn{1}{l|}{0.835} &
  \multicolumn{1}{l|}{0.771} &
  0.838 &
  \multicolumn{1}{l|}{0.667} &
  \multicolumn{1}{l|}{0.588} &
  0.708 \\ \hline
Kidney Disease &
  \multicolumn{1}{l|}{1.000} &
  \multicolumn{1}{l|}{0.976} &
  0.988 &
  \multicolumn{1}{l|}{0.938} &
  \multicolumn{1}{l|}{0.988} &
  0.964 &
  \multicolumn{1}{l|}{0.646} &
  \multicolumn{1}{l|}{0.754} &
  0.806 \\ \hline
Liver Disease &
  \multicolumn{1}{l|}{0.827} &
  \multicolumn{1}{l|}{0.834} &
  0.820 &
  \multicolumn{1}{l|}{0.819} &
  \multicolumn{1}{l|}{0.803} &
  0.817 &
  \multicolumn{1}{l|}{0.717} &
  \multicolumn{1}{l|}{0.804} &
  0.795 \\ \hline
Maternal Risk &
  \multicolumn{1}{l|}{0.855} &
  \multicolumn{1}{l|}{0.476} &
  0.837 &
  \multicolumn{1}{l|}{0.803} &
  \multicolumn{1}{l|}{0.812} &
  0.780 &
  \multicolumn{1}{l|}{0.739} &
  \multicolumn{1}{l|}{0.735} &
  0.689 \\ \hline
Parkinsons &
  \multicolumn{1}{l|}{0.929} &
  \multicolumn{1}{l|}{0.860} &
  0.935 &
  \multicolumn{1}{l|}{0.894} &
  \multicolumn{1}{l|}{0.836} &
  0.875 &
  \multicolumn{1}{l|}{0.707} &
  \multicolumn{1}{l|}{0.818} &
  0.732 \\ \hline
Parkinsons Biom. &
  \multicolumn{1}{l|}{0.203} &
  \multicolumn{1}{l|}{0.476} &
  0.167 &
  \multicolumn{1}{l|}{0.337} &
  \multicolumn{1}{l|}{0.265} &
  0.192 &
  \multicolumn{1}{l|}{0.133} &
  \multicolumn{1}{l|}{0.233} &
  0.182 \\ \hline
Spine &
  \multicolumn{1}{l|}{0.933} &
  \multicolumn{1}{l|}{0.652} &
  0.954 &
  \multicolumn{1}{l|}{0.932} &
  \multicolumn{1}{l|}{0.907} &
  0.926 &
  \multicolumn{1}{l|}{0.775} &
  \multicolumn{1}{l|}{0.818} &
  0.742 \\ \hline
Stroke &
  \multicolumn{1}{l|}{0.239} &
  \multicolumn{1}{l|}{0.474} &
  0.244 &
  \multicolumn{1}{l|}{0.225} &
  \multicolumn{1}{l|}{0.255} &
  0.153 &
  \multicolumn{1}{l|}{0.229} &
  \multicolumn{1}{l|}{0.255} &
  0.208 \\ \hline
\end{tabular}
\end{table}

Table \ref{tab5} reports the statistical significance results for the comparison of the Auto-PU systems in terms of F-measure, precision and recall. In Table \ref{tab5}, for each combination of a performance measure (F-measure, precision, recall) and a \(\delta\) value (\(\delta\)= 20\%, 40\%, 60\%), this table has one row for each pair of Auto-PU systems being compared, and for each such pairwise comparison, the table reports the average (avg.) rank of each of the two methods being compared and the corresponding p-value and adjusted \(\alpha\) value (significance level), based on the Holm correction \citep{demvsar2006statistical}. The better (lower) avg. rank in each cell is shown in boldface. For example, in the cell for F-measure, \(\delta\) = 20\%, Compared Systems GA-1 vs BO-1, the average ranks for those two systems are 1.52 and 1.48, respectively. Hence, BO-1 was the winner, but the p-value (0.952) was greater than the adjusted \(\alpha\), so this result was not statistically significant. The following discussion of results will focus mainly on the F-measure, the most important measure in Table \ref{tab5}, as mentioned earlier.

\begin{table}[htbp]
\caption{Results of Wilcoxon signed-rank tests with Holm correction for multiple hypothesis when comparing each pair of Auto-PU systems (with the base search space) regarding F-measure, Precision and Recall, for the 3 \(\delta\) values.}\label{tab5}
\centering
\small
\setlength{\tabcolsep}{3pt}
\begin{tabular}{|l|l|lll|lll|lll|}
\hline
\multirow{2}{*}{\(\delta\) (\%)} &
  \multirow{2}{*}{\begin{tabular}[c]{@{}l@{}}Compared \\ systems\end{tabular}} &
  \multicolumn{3}{c|}{F-measure} &
  \multicolumn{3}{c|}{Precision} &
  \multicolumn{3}{c|}{Recall} \\ \cline{3-11} 
 &
   &
  \multicolumn{1}{l|}{\begin{tabular}[c]{@{}l@{}}Avg \\ ranks\end{tabular}} &
  \multicolumn{1}{l|}{p-value} &
  \(\alpha\) &
  \multicolumn{1}{l|}{\begin{tabular}[c]{@{}l@{}}Avg   \\ ranks\end{tabular}} &
  \multicolumn{1}{l|}{p-value} &
  \(\alpha\) &
  \multicolumn{1}{l|}{\begin{tabular}[c]{@{}l@{}}Avg \\ ranks\end{tabular}} &
  \multicolumn{1}{l|}{p-value} &
  \(\alpha\) \\ \hline
\multirow{3}{*}{20\%} &
  \begin{tabular}[c]{@{}l@{}}GA-1 vs \\ BO-1\end{tabular} &
  \multicolumn{1}{l|}{\begin{tabular}[c]{@{}l@{}}1.52 vs \\ \textbf{1.48}\end{tabular}} &
  \multicolumn{1}{l|}{0.952} &
  0.05 &
  \multicolumn{1}{l|}{\begin{tabular}[c]{@{}l@{}}1.55 vs \\ \textbf{1.45}\end{tabular}} &
  \multicolumn{1}{l|}{0.983} &
  0.05 &
  \multicolumn{1}{l|}{\begin{tabular}[c]{@{}l@{}}\textbf{1.4} vs \\ 1.6\end{tabular}} &
  \multicolumn{1}{l|}{0.114} &
  0.017 \\ \cline{2-11} 
 &
  \begin{tabular}[c]{@{}l@{}}GA-1 vs \\ EBO-1\end{tabular} &
  \multicolumn{1}{l|}{\begin{tabular}[c]{@{}l@{}}1.5 vs \\ 1.5\end{tabular}} &
  \multicolumn{1}{l|}{0.784} &
  0.025 &
  \multicolumn{1}{l|}{\begin{tabular}[c]{@{}l@{}}1.52 vs \\ \textbf{1.48}\end{tabular}} &
  \multicolumn{1}{l|}{0.446} &
  0.017 &
  \multicolumn{1}{l|}{\begin{tabular}[c]{@{}l@{}}\textbf{1.38} vs \\ 1.62\end{tabular}} &
  \multicolumn{1}{l|}{0.178} &
  0.025 \\ \cline{2-11} 
 &
  \begin{tabular}[c]{@{}l@{}}BO-1 vs \\ EBO-1\end{tabular} &
  \multicolumn{1}{l|}{\begin{tabular}[c]{@{}l@{}}1.52 vs \\ \textbf{1.48}\end{tabular}} &
  \multicolumn{1}{l|}{0.658} &
  0.017 &
  \multicolumn{1}{l|}{\begin{tabular}[c]{@{}l@{}}1.55 vs \\ \textbf{1.45}\end{tabular}} &
  \multicolumn{1}{l|}{0.968} &
  0.025 &
  \multicolumn{1}{l|}{\begin{tabular}[c]{@{}l@{}}1.55 vs \\ \textbf{1.45}\end{tabular}} &
  \multicolumn{1}{l|}{0.983} &
  0.05 \\ \hline
\multirow{3}{*}{40\%} &
  \begin{tabular}[c]{@{}l@{}}GA-1 vs \\ BO-1\end{tabular} &
  \multicolumn{1}{l|}{\begin{tabular}[c]{@{}l@{}}1.62 vs \\ \textbf{1.38}\end{tabular}} &
  \multicolumn{1}{l|}{0.334} &
  0.025 &
  \multicolumn{1}{l|}{\begin{tabular}[c]{@{}l@{}}1.55 vs \\ \textbf{1.45}\end{tabular}} &
  \multicolumn{1}{l|}{0.601} &
  0.025 &
  \multicolumn{1}{l|}{\begin{tabular}[c]{@{}l@{}}1.58 vs \\ \textbf{1.42}\end{tabular}} &
  \multicolumn{1}{l|}{0.398} &
  0.025 \\ \cline{2-11} 
 &
  \begin{tabular}[c]{@{}l@{}}GA-1 vs \\ EBO-1\end{tabular} &
  \multicolumn{1}{l|}{\begin{tabular}[c]{@{}l@{}}1.6 vs \\ \textbf{1.4}\end{tabular}} &
  \multicolumn{1}{l|}{0.245} &
  0.017 &
  \multicolumn{1}{l|}{\begin{tabular}[c]{@{}l@{}}\textbf{1.45} vs \\ 1.55\end{tabular}} &
  \multicolumn{1}{l|}{0.828} &
  0.05 &
  \multicolumn{1}{l|}{\begin{tabular}[c]{@{}l@{}}1.58 vs \\ \textbf{1.42}\end{tabular}} &
  \multicolumn{1}{l|}{0.381} &
  0.017 \\ \cline{2-11} 
 &
  \begin{tabular}[c]{@{}l@{}}BO-1 vs \\ EBO-1\end{tabular} &
  \multicolumn{1}{l|}{\begin{tabular}[c]{@{}l@{}}1.55 vs \\ \textbf{1.45}\end{tabular}} &
  \multicolumn{1}{l|}{1.000} &
  0.05 &
  \multicolumn{1}{l|}{\begin{tabular}[c]{@{}l@{}}\textbf{1.45} vs \\ 1.55\end{tabular}} &
  \multicolumn{1}{l|}{0.387} &
  0.017 &
  \multicolumn{1}{l|}{\begin{tabular}[c]{@{}l@{}}1.5 vs \\ 1.5\end{tabular}} &
  \multicolumn{1}{l|}{0.557} &
  0.05 \\ \hline
\multirow{3}{*}{60\%} &
  \begin{tabular}[c]{@{}l@{}}GA-1 vs \\ BO-1\end{tabular} &
  \multicolumn{1}{l|}{\begin{tabular}[c]{@{}l@{}}\textbf{1.25} vs \\ 1.75\end{tabular}} &
  \multicolumn{1}{l|}{0.177} &
  0.017 &
  \multicolumn{1}{l|}{\begin{tabular}[c]{@{}l@{}}\textbf{1.48} vs \\ 1.52\end{tabular}} &
  \multicolumn{1}{l|}{0.513} &
  0.025 &
  \multicolumn{1}{l|}{\begin{tabular}[c]{@{}l@{}}\textbf{1.25} vs \\ 1.75\end{tabular}} &
  \multicolumn{1}{l|}{0.064} &
  0.017 \\ \cline{2-11} 
 &
  \begin{tabular}[c]{@{}l@{}}GA-1 vs \\ EBO-1\end{tabular} &
  \multicolumn{1}{l|}{\begin{tabular}[c]{@{}l@{}}\textbf{1.45} vs \\ 1.55\end{tabular}} &
  \multicolumn{1}{l|}{0.312} &
  0.025 &
  \multicolumn{1}{l|}{\begin{tabular}[c]{@{}l@{}}1.5 vs \\ 1.5\end{tabular}} &
  \multicolumn{1}{l|}{0.812} &
  0.05 &
  \multicolumn{1}{l|}{\begin{tabular}[c]{@{}l@{}}\textbf{1.45} vs \\ 1.55\end{tabular}} &
  \multicolumn{1}{l|}{0.388} &
  0.025 \\ \cline{2-11} 
 &
  \begin{tabular}[c]{@{}l@{}}BO-1 vs \\ EBO-1\end{tabular} &
  \multicolumn{1}{l|}{\begin{tabular}[c]{@{}l@{}}1.6 vs \\ \textbf{1.4}\end{tabular}} &
  \multicolumn{1}{l|}{0.674} &
  0.05 &
  \multicolumn{1}{l|}{\begin{tabular}[c]{@{}l@{}}\textbf{1.38} vs \\ 1.62\end{tabular}} &
  \multicolumn{1}{l|}{0.184} &
  0.017 &
  \multicolumn{1}{l|}{\begin{tabular}[c]{@{}l@{}}1.62 vs \\ \textbf{1.38}\end{tabular}} &
  \multicolumn{1}{l|}{0.629} &
  0.05 \\ \hline
\end{tabular}
\end{table}

The results in Table \ref{tab5} show very little difference in performance between the methods for \(\delta\) = 20\%, with no statistically significant results. For the F-measure results, in the pairwise comparisons, EBO-1 ties with GA-1 and slightly outperforms BO-1; and BO-1 slightly outperformed GA-1. The results for \(\delta\) = 40\% show a slightly larger (but still small) range in performance in regard to F-measure; with EBO-1 slightly outperforming both GA-1 and BO-1, whilst BO-1 outperforms GA-1. However, again there are no statistically significant results (for any of the three measures) for \(\delta\) = 40\%. For \(\delta\) = 60\%, regarding the F-measure results, GA-1 substantially outperformed BO-1 (avg. rank \textbf{1.25} vs 1.75) and slightly outperformed EBO-1; whilst EBO-1 slightly outperformed BO-1. However, again there are no statistically significant results for \(\delta\) = 60\%.

Based on these results, it can be argued that overall EBO-1 performed best for \(\delta\) = 20\% and \(\delta\) = 40\%, whilst GA-1 performed best for \(\delta\) = 60\%; but no statistical significance was achieved in any experiment, and the performance gains of EBO-1 and GA-1 were in general marginal – except that GA-1 had a substantially better avg. rank than BO-1 for \(\delta\) = 60\%.  

\begin{figure}[htbp]
    \centering
    \includegraphics[width=0.6\textwidth]{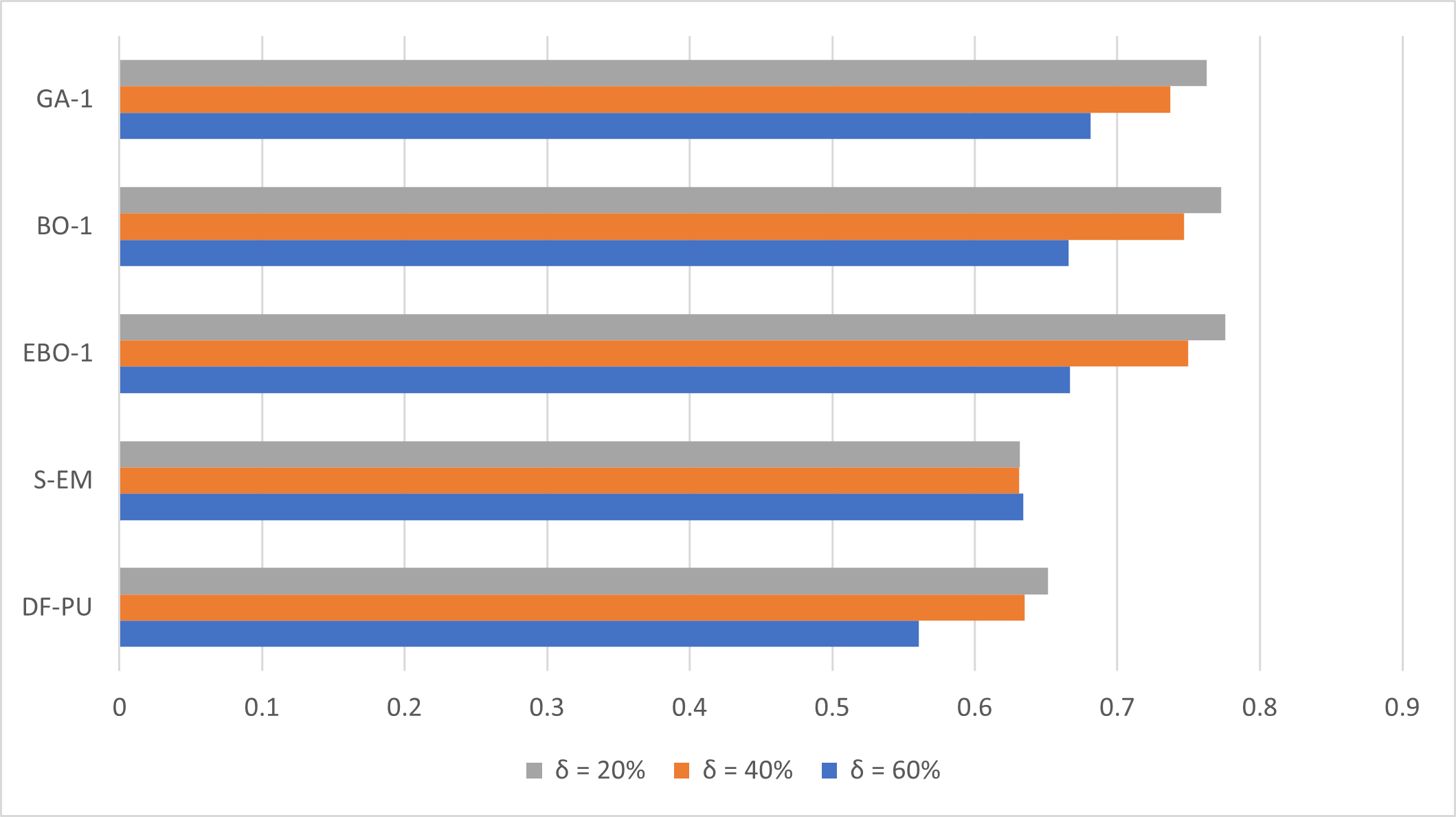}
    \caption{Average F-measure results comparison for three versions of Auto-PU utilising the base search space and the PU learning baselines across the three values of \(\delta\).}
    \label{fig3}
\end{figure}

Figure \ref{fig3} shows graphically how the F-measure values achieved by the three Auto-PU systems and the two baseline PU learning methods (DF-PU and S-EM) vary across the three \(\delta\) values. The results of the baseline methods will be discussed in the next Subsection.

Focusing for now on the results for GA-1, BO-1 and EBO-1 in this figure, it is clear that, for all the three systems, their achieved F-measure values decrease with an increase in the \(\delta\) value, i.e. they all achieved their best F-measure value when \(\delta\) = 20\% and their worst F-measure value when \(\delta\) = 60\%. This is not surprising, considering that \(\delta\) represents the percentage of positive instances hidden in the unlabelled set; and intuitively, the larger this percentage, the more difficult the PU learning problem is. The decrease in F-measure values is relatively small when \(\delta\) is increased from 20\% to 40\%, but there is a more substantial F-measure value decrease when \(\delta\) is increased to 60\%, for all the three Auto-PU systems.

\begin{table}[htbp]
\caption{Linear (Pearson’s) correlation coefficient value between the F-measure and the percentage of positive examples in the original dataset (before hiding some positive examples in the unlabelled set) for each combination of an Auto-PU system (with base search space) or PU method and a \(\delta\) value}\label{tab6}
\centering
\small
\begin{tabular}{|l|c|c|c|}
\hline
\begin{tabular}[c]{@{}l@{}}Auto-PU system \\ or PU method\end{tabular} & \(\delta\) = 20\% & \(\delta\) = 40\% & \(\delta\) = 60\% \\ \hline
EBO-1 & 0.440 & 0.469 & 0.460 \\ \hline
BO-1  & 0.398 & 0.360 & 0.498 \\ \hline
GA-1  & 0.333 & 0.385 & 0.504 \\ \hline
DF-PU & 0.381 & 0.504 & 0.445 \\ \hline
S-EM  & 0.690 & 0.631 & 0.686 \\ \hline
\end{tabular}
\end{table}

In order to further analyse the results, Table \ref{tab6} shows the values of Pearson’s linear correlation coefficient between the F-measure values achieved by each Auto-PU system or baseline PU learning method and percentages of positive examples in the original dataset, for each \(\delta\) value. For all systems or methods, across all \(\delta\) values, there is a positive correlation between these two factors, indicating that in general, the larger the percentage of positive-class instances, the higher the F-measure value, probable due to the smaller degree of class imbalance in the dataset. To simplify the analysis of these correlations, we categorise the positive correlations as defined by \cite{schober2018correlation}: 0.00-0.10 negligible correlation, 0.10-0.39 weak correlation, 0.40-0.69 moderate correlation, 0.70-0.89 strong correlation, 0.90-1.00 very strong correlation. 

Of the three Auto-PU systems, EBO-1 shows the strongest positive correlation overall, achieving a moderate correlation for each \(\delta\) value. BO-1 and GA-1 both show weak correlations for \(\delta\) = 20\% and \(\delta\) = 40\%, but both show moderate correlation for \(\delta\) = 60\%. Whilst the correlation for EBO-1 is relatively stable across \(\delta\) values, BO-1 and GA-1 both show a substantial increase for \(\delta\) = 60\%. 
The correlations for the two baseline PU learning methods will be discussed in the next subsection.

\subsubsection{Comparing the Auto-PU Systems against two baseline PU methods}

Table \ref{tab7} reports the F-measure values achieved by the baseline PU learning methods DF-PU and S-EM (see Section 2.1), whilst Table \ref{tab8} reports the the statistical significance results for F-measure, precision and recall when comparing the Auto-PU systems against the two baseline PU learning methods. Table \ref{tab8} has three parts, one part for each \(\delta\) value (shown in the first column). In each part, the table shows the statistical results for 6 pairwise comparisons, by comparing each of the three Auto-PU systems against each of the two baseline PU learning methods. The statistical results for each pairwise comparison follow the same format as described earlier for Table \ref{tab5}. In each comparison, the better (lower) avg. rank in each cell is shown in boldface; and a p-value is shown in boldface if it is significant (i.e., smaller than the adjusted \(\alpha\)). For example, in the first row of results in Table \ref{tab8}, for \(\delta\)= 20\%, comparing GA-1 vs DF-PU, regarding F-measure, the average ranks of GA-1 and DF-PU were 1.05 and 1.95 respectively, and this result was statistically significant. The following discussion of results will focus mainly on the F-measure, the most important measure in Table \ref{tab8}, whilst precision and recall results are reported for completeness.

\begin{table}[]
\caption{F-measure results of the baseline PU learning methods.}
\label{tab7}
\centering
\small
\begin{tabular}{|l|ll|ll|ll|}
\hline
\multicolumn{1}{|c|}{\multirow{2}{*}{Dataset}} & \multicolumn{2}{c|}{\(\delta\) = 20\%} & \multicolumn{2}{c|}{\(\delta\) = 40\%} & \multicolumn{2}{c|}{\(\delta\) = 60\%} \\ \cline{2-7} 
\multicolumn{1}{|c|}{}                         & \multicolumn{1}{l|}{DF-PU}   & S-EM    & \multicolumn{1}{l|}{DF-PU}   & S-EM    & \multicolumn{1}{l|}{DF-PU}   & S-EM    \\ \hline
Alzheimer’s                                    & \multicolumn{1}{l|}{0.519}   & 0.385   & \multicolumn{1}{l|}{0.396}   & 0.489   & \multicolumn{1}{l|}{0.462}   & 0.389   \\ \hline
Autism                                         & \multicolumn{1}{l|}{0.766}   & 0.823   & \multicolumn{1}{l|}{0.639}   & 0.824   & \multicolumn{1}{l|}{0.557}   & 0.838   \\ \hline
Breast cancer Coi.                             & \multicolumn{1}{l|}{0.680}   & 0.896   & \multicolumn{1}{l|}{0.894}   & 0.903   & \multicolumn{1}{l|}{0.618}   & 0.903   \\ \hline
Breast cancer Wis.                             & \multicolumn{1}{l|}{0.884}   & 0.892   & \multicolumn{1}{l|}{0.700}   & 0.893   & \multicolumn{1}{l|}{0.473}   & 0.892   \\ \hline
Breast cancer mut.                             & \multicolumn{1}{l|}{0.500}   & 0.711   & \multicolumn{1}{l|}{0.500}   & 0.690   & \multicolumn{1}{l|}{0.500}   & 0.702   \\ \hline
Cervical cancer                                & \multicolumn{1}{l|}{0.688}   & 0.054   & \multicolumn{1}{l|}{0.667}   & 0.053   & \multicolumn{1}{l|}{0.650}   & 0.046   \\ \hline
Cirrhosis                                      & \multicolumn{1}{l|}{0.458}   & 0.438   & \multicolumn{1}{l|}{0.389}   & 0.442   & \multicolumn{1}{l|}{0.364}   & 0.465   \\ \hline
Dermatology                                    & \multicolumn{1}{l|}{0.872}   & 0.718   & \multicolumn{1}{l|}{0.712}   & 0.718   & \multicolumn{1}{l|}{0.860}   & 0.719   \\ \hline
PI Diabetes                                    & \multicolumn{1}{l|}{0.566}   & 0.536   & \multicolumn{1}{l|}{0.582}   & 0.535   & \multicolumn{1}{l|}{0.584}   & 0.575   \\ \hline
ES Diabetes                                    & \multicolumn{1}{l|}{0.751}   & 0.796   & \multicolumn{1}{l|}{0.726}   & 0.863   & \multicolumn{1}{l|}{0.700}   & 0.783   \\ \hline
Heart Disease                                  & \multicolumn{1}{l|}{0.719}   & 0.838   & \multicolumn{1}{l|}{0.701}   & 0.838   & \multicolumn{1}{l|}{0.730}   & 0.828   \\ \hline
Heart Failure                                  & \multicolumn{1}{l|}{0.590}   & 0.490   & \multicolumn{1}{l|}{0.586}   & 0.514   & \multicolumn{1}{l|}{0.457}   & 0.584   \\ \hline
Hepatitis C                                    & \multicolumn{1}{l|}{0.916}   & 0.701   & \multicolumn{1}{l|}{0.804}   & 0.708   & \multicolumn{1}{l|}{0.683}   & 0.661   \\ \hline
Kidney Disease                                 & \multicolumn{1}{l|}{0.874}   & 1.000   & \multicolumn{1}{l|}{1.000}   & 1.000   & \multicolumn{1}{l|}{0.607}   & 0.951   \\ \hline
Liver Disease                                  & \multicolumn{1}{l|}{0.806}   & 0.834   & \multicolumn{1}{l|}{0.829}   & 0.726   & \multicolumn{1}{l|}{0.753}   & 0.798   \\ \hline
Maternal Risk                                  & \multicolumn{1}{l|}{0.459}   & 0.457   & \multicolumn{1}{l|}{0.474}   & 0.444   & \multicolumn{1}{l|}{0.418}   & 0.499   \\ \hline
Parkinsons                                     & \multicolumn{1}{l|}{0.857}   & 0.847   & \multicolumn{1}{l|}{0.857}   & 0.751   & \multicolumn{1}{l|}{0.799}   & 0.767   \\ \hline
Parkinsons Biom.                               & \multicolumn{1}{l|}{0.313}   & 0.308   & \multicolumn{1}{l|}{0.375}   & 0.276   & \multicolumn{1}{l|}{0.065}   & 0.331   \\ \hline
Spine                                          & \multicolumn{1}{l|}{0.651}   & 0.806   & \multicolumn{1}{l|}{0.654}   & 0.852   & \multicolumn{1}{l|}{0.723}   & 0.840   \\ \hline
Stroke                                         & \multicolumn{1}{l|}{0.155}   & 0.102   & \multicolumn{1}{l|}{0.214}   & 0.102   & \multicolumn{1}{l|}{0.207}   & 0.103   \\ \hline
\end{tabular}
\end{table}

\begin{table}[]
\caption{Results of Wilcoxon signed-rank tests with Holm correction for multiple hypothesis when comparing each of GA-1, BO-1 and EBO-1 (with the base search space) against each of two baseline PU learning methods (DF-PU and S-EM) regarding F-measure, Precision and Recall, for the three \(\delta\) values.}
\label{tab8}
\centering
\small
\setlength{\tabcolsep}{3pt}
\begin{tabular}{|c|l|lll|lll|lll|}
\hline
\multicolumn{1}{|l|}{\multirow{2}{*}{\(\delta\) (\%)}} & \multirow{2}{*}{\begin{tabular}[c]{@{}l@{}}Compared \\ systems\end{tabular}} & \multicolumn{3}{c|}{F-measure}                                                                                                                                                  & \multicolumn{3}{c|}{Precision}                                                                                                                                                  & \multicolumn{3}{c|}{Recall}                                                                                                                                                     \\ \cline{3-11} 
\multicolumn{1}{|l|}{}                               &                                                                              & \multicolumn{1}{l|}{Ranking}                                                 & \multicolumn{1}{l|}{P-value}        & \begin{tabular}[c]{@{}l@{}}Adj. \\ \(\alpha\)\end{tabular} & \multicolumn{1}{l|}{Ranking}                                                 & \multicolumn{1}{l|}{P-value}        & \begin{tabular}[c]{@{}l@{}}Adj. \\ \(\alpha\)\end{tabular} & \multicolumn{1}{l|}{Ranking}                                                 & \multicolumn{1}{l|}{P-value}        & \begin{tabular}[c]{@{}l@{}}Adj. \\ \(\alpha\)\end{tabular} \\ \hline
\multirow{6}{*}{20}                                  & \begin{tabular}[c]{@{}l@{}}GA-1 vs \\ DF-PU\end{tabular}                     & \multicolumn{1}{l|}{\begin{tabular}[c]{@{}l@{}}\textbf{1.05} vs \\ 1.95\end{tabular}} & \multicolumn{1}{l|}{\textbf{1E-05}} & 0.025                                                      & \multicolumn{1}{l|}{\begin{tabular}[c]{@{}l@{}}\textbf{1.2} vs \\ 1.8\end{tabular}}   & \multicolumn{1}{l|}{\textbf{0.001}} & 0.025                                                      & \multicolumn{1}{l|}{\begin{tabular}[c]{@{}l@{}}\textbf{1.38 }vs \\ 1.62\end{tabular}} & \multicolumn{1}{l|}{0.494}          & 0.05                                                       \\ \cline{2-11} 
                                                     & \begin{tabular}[c]{@{}l@{}}GA-1 vs \\ S-EM\end{tabular}                      & \multicolumn{1}{l|}{\begin{tabular}[c]{@{}l@{}}\textbf{1.15} vs \\ 1.85\end{tabular}} & \multicolumn{1}{l|}{\textbf{0.006}} & 0.05                                                       & \multicolumn{1}{l|}{\begin{tabular}[c]{@{}l@{}}\textbf{1.18} vs \\ 1.82\end{tabular}} & \multicolumn{1}{l|}{\textbf{0.002}} & 0.05                                                       & \multicolumn{1}{l|}{\begin{tabular}[c]{@{}l@{}}1.85 vs \\ \textbf{1.15}\end{tabular}} & \multicolumn{1}{l|}{\textbf{0.011}} & 0.025                                                      \\ \cline{2-11} 
                                                     & \begin{tabular}[c]{@{}l@{}}BO-1 vs \\ DF-PU\end{tabular}                     & \multicolumn{1}{l|}{\begin{tabular}[c]{@{}l@{}}\textbf{1.15} vs \\ 1.85\end{tabular}} & \multicolumn{1}{l|}{\textbf{3E-04}} & 0.025                                                      & \multicolumn{1}{l|}{\begin{tabular}[c]{@{}l@{}}\textbf{1.25} vs \\ 1.75\end{tabular}} & \multicolumn{1}{l|}{\textbf{0.002}} & 0.025                                                      & \multicolumn{1}{l|}{\begin{tabular}[c]{@{}l@{}}\textbf{1.4} vs \\ 1.6\end{tabular}}   & \multicolumn{1}{l|}{0.189}          & 0.025                                                      \\ \cline{2-11} 
                                                     & \begin{tabular}[c]{@{}l@{}}BO-1 vs \\ S-EM\end{tabular}                      & \multicolumn{1}{l|}{\begin{tabular}[c]{@{}l@{}}\textbf{1.2} vs \\ 1.8\end{tabular}}   & \multicolumn{1}{l|}{\textbf{0.003}} & 0.05                                                       & \multicolumn{1}{l|}{\begin{tabular}[c]{@{}l@{}}\textbf{1.22} vs \\ 1.78\end{tabular}} & \multicolumn{1}{l|}{\textbf{0.005}} & 0.05                                                       & \multicolumn{1}{l|}{\begin{tabular}[c]{@{}l@{}}1.68 vs \\ \textbf{1.32}\end{tabular}} & \multicolumn{1}{l|}{0.266}          & 0.05                                                       \\ \cline{2-11} 
                                                     & \begin{tabular}[c]{@{}l@{}}EBO-1 vs \\ DF-PU\end{tabular}                    & \multicolumn{1}{l|}{\begin{tabular}[c]{@{}l@{}}\textbf{1.05} vs \\ 1.95\end{tabular}} & \multicolumn{1}{l|}{\textbf{6E-05}} & 0.025                                                      & \multicolumn{1}{l|}{\begin{tabular}[c]{@{}l@{}}\textbf{1.1} vs \\ 1.9\end{tabular}}   & \multicolumn{1}{l|}{\textbf{1E-05}} & 0.025                                                      & \multicolumn{1}{l|}{\begin{tabular}[c]{@{}l@{}}\textbf{1.38} vs \\ 1.62\end{tabular}} & \multicolumn{1}{l|}{0.494}          & 0.025                                                      \\ \cline{2-11} 
                                                     & \begin{tabular}[c]{@{}l@{}}EBO-1 vs \\ S-EM\end{tabular}                     & \multicolumn{1}{l|}{\begin{tabular}[c]{@{}l@{}}\textbf{1.22} vs \\ 1.78\end{tabular}} & \multicolumn{1}{l|}{\textbf{0.002}} & 0.05                                                       & \multicolumn{1}{l|}{\begin{tabular}[c]{@{}l@{}}\textbf{1.08} vs \\ 1.92\end{tabular}} & \multicolumn{1}{l|}{\textbf{2E-04}} & 0.05                                                       & \multicolumn{1}{l|}{\begin{tabular}[c]{@{}l@{}}1.78 vs \\ \textbf{1.22}\end{tabular}} & \multicolumn{1}{l|}{\textbf{0.018}} & 0.05                                                       \\ \hline
\multirow{6}{*}{40}                                  & \begin{tabular}[c]{@{}l@{}}GA-1 vs \\ DF-PU\end{tabular}                     & \multicolumn{1}{l|}{\begin{tabular}[c]{@{}l@{}}\textbf{1.3} vs \\ 1.7\end{tabular}}   & \multicolumn{1}{l|}{\textbf{0.007}} & 0.05                                                       & \multicolumn{1}{l|}{\begin{tabular}[c]{@{}l@{}}\textbf{1.12} vs \\ 1.88\end{tabular}} & \multicolumn{1}{l|}{\textbf{0.001}} & 0.05                                                       & \multicolumn{1}{l|}{\begin{tabular}[c]{@{}l@{}}1.55 vs \\ \textbf{1.45}\end{tabular}} & \multicolumn{1}{l|}{0.841}          & 0.05                                                       \\ \cline{2-11} 
                                                     & \begin{tabular}[c]{@{}l@{}}GA-1 vs \\ S-EM\end{tabular}                      & \multicolumn{1}{l|}{\begin{tabular}[c]{@{}l@{}}\textbf{1.2} vs \\ 1.8\end{tabular}}   & \multicolumn{1}{l|}{\textbf{0.002}} & 0.025                                                      & \multicolumn{1}{l|}{\begin{tabular}[c]{@{}l@{}}\textbf{1.12} vs \\ 1.88\end{tabular}} & \multicolumn{1}{l|}{\textbf{0.002}} & 0.025                                                      & \multicolumn{1}{l|}{\begin{tabular}[c]{@{}l@{}}1.75 vs \\ \textbf{1.25}\end{tabular}} & \multicolumn{1}{l|}{\textbf{0.011}} & 0.025                                                      \\ \cline{2-11} 
                                                     & \begin{tabular}[c]{@{}l@{}}BO-1 vs \\ DF-PU\end{tabular}                     & \multicolumn{1}{l|}{\begin{tabular}[c]{@{}l@{}}\textbf{1.25} vs \\ 1.75\end{tabular}} & \multicolumn{1}{l|}{\textbf{0.008}} & 0.05                                                       & \multicolumn{1}{l|}{\begin{tabular}[c]{@{}l@{}}\textbf{1.12} vs \\ 1.88\end{tabular}} & \multicolumn{1}{l|}{\textbf{0.001}} & 0.025                                                      & \multicolumn{1}{l|}{\begin{tabular}[c]{@{}l@{}}1.58 vs \\ \textbf{1.42}\end{tabular}} & \multicolumn{1}{l|}{0.629}          & 0.05                                                       \\ \cline{2-11} 
                                                     & \begin{tabular}[c]{@{}l@{}}BO-1 vs \\ S-EM\end{tabular}                      & \multicolumn{1}{l|}{\begin{tabular}[c]{@{}l@{}}\textbf{1.2} vs \\ 1.8\end{tabular}}   & \multicolumn{1}{l|}{\textbf{0.003}} & 0.025                                                      & \multicolumn{1}{l|}{\begin{tabular}[c]{@{}l@{}}\textbf{1.12} vs \\ 1.88\end{tabular}} & \multicolumn{1}{l|}{\textbf{0.002}} & 0.05                                                       & \multicolumn{1}{l|}{\begin{tabular}[c]{@{}l@{}}1.8 vs \\ \textbf{1.2}\end{tabular}}   & \multicolumn{1}{l|}{\textbf{0.003}} & 0.025                                                      \\ \cline{2-11} 
                                                     & \begin{tabular}[c]{@{}l@{}}EBO-1 vs \\ DF-PU\end{tabular}                    & \multicolumn{1}{l|}{\begin{tabular}[c]{@{}l@{}}\textbf{1.15} vs \\ 1.85\end{tabular}} & \multicolumn{1}{l|}{\textbf{2E-04}} & 0.025                                                      & \multicolumn{1}{l|}{\begin{tabular}[c]{@{}l@{}}\textbf{1.08} vs \\ 1.92\end{tabular}} & \multicolumn{1}{l|}{\textbf{2E-04}} & 0.025                                                      & \multicolumn{1}{l|}{\begin{tabular}[c]{@{}l@{}}1.5 vs \\ 1.5\end{tabular}}   & \multicolumn{1}{l|}{0.777}          & 0.05                                                       \\ \cline{2-11} 
                                                     & \begin{tabular}[c]{@{}l@{}}EBO-1 vs \\ S-EM\end{tabular}                     & \multicolumn{1}{l|}{\begin{tabular}[c]{@{}l@{}}\textbf{1.2} vs \\ 1.8\end{tabular}}   & \multicolumn{1}{l|}{\textbf{4E-04}} & 0.05                                                       & \multicolumn{1}{l|}{\begin{tabular}[c]{@{}l@{}}\textbf{1.08} vs \\ 1.92\end{tabular}} & \multicolumn{1}{l|}{\textbf{3E-04}} & 0.05                                                       & \multicolumn{1}{l|}{\begin{tabular}[c]{@{}l@{}}1.7 vs \\ \textbf{1.3}\end{tabular}}   & \multicolumn{1}{l|}{0.030}          & 0.025                                                      \\ \hline
\multirow{6}{*}{60}                                  & \begin{tabular}[c]{@{}l@{}}GA-1 vs \\ DF-PU\end{tabular}                     & \multicolumn{1}{l|}{\begin{tabular}[c]{@{}l@{}}\textbf{1.2} vs \\ 1.8\end{tabular}}   & \multicolumn{1}{l|}{\textbf{0.003}} & 0.025                                                      & \multicolumn{1}{l|}{\begin{tabular}[c]{@{}l@{}}\textbf{1.1} vs \\ 1.9\end{tabular}}   & \multicolumn{1}{l|}{\textbf{0.001}} & 0.05                                                       & \multicolumn{1}{l|}{\begin{tabular}[c]{@{}l@{}}1.55 vs \\ \textbf{1.45}\end{tabular}} & \multicolumn{1}{l|}{0.571}          & 0.05                                                       \\ \cline{2-11} 
                                                     & \begin{tabular}[c]{@{}l@{}}GA-1 vs \\ S-EM\end{tabular}                      & \multicolumn{1}{l|}{\begin{tabular}[c]{@{}l@{}}\textbf{1.35} vs \\ 1.65\end{tabular}} & \multicolumn{1}{l|}{0.216}          & 0.05                                                       & \multicolumn{1}{l|}{\begin{tabular}[c]{@{}l@{}}\textbf{1.22} vs \\ 1.78\end{tabular}} & \multicolumn{1}{l|}{\textbf{0.001}} & 0.025                                                      & \multicolumn{1}{l|}{\begin{tabular}[c]{@{}l@{}}2.0 vs \\ \textbf{1.0}\end{tabular}}   & \multicolumn{1}{l|}{\textbf{2E-06}} & 0.025                                                      \\ \cline{2-11} 
                                                     & \begin{tabular}[c]{@{}l@{}}BO-1 vs \\ DF-PU\end{tabular}                     & \multicolumn{1}{l|}{\begin{tabular}[c]{@{}l@{}}\textbf{1.25} vs \\ 1.75\end{tabular}} & \multicolumn{1}{l|}{\textbf{0.011}} & 0.025                                                      & \multicolumn{1}{l|}{\begin{tabular}[c]{@{}l@{}}\textbf{1.1} vs \\ 1.9\end{tabular}}   & \multicolumn{1}{l|}{\textbf{0.001}} & 0.025                                                      & \multicolumn{1}{l|}{\begin{tabular}[c]{@{}l@{}}1.52 vs \\ \textbf{1.48}\end{tabular}} & \multicolumn{1}{l|}{0.748}          & 0.05                                                       \\ \cline{2-11} 
                                                     & \begin{tabular}[c]{@{}l@{}}BO-1 vs \\ S-EM\end{tabular}                      & \multicolumn{1}{l|}{\begin{tabular}[c]{@{}l@{}}\textbf{1.35} vs \\ 1.65\end{tabular}} & \multicolumn{1}{l|}{0.409}          & 0.05                                                       & \multicolumn{1}{l|}{\begin{tabular}[c]{@{}l@{}}\textbf{1.12} vs \\ 1.88\end{tabular}} & \multicolumn{1}{l|}{\textbf{0.001}} & 0.05                                                       & \multicolumn{1}{l|}{\begin{tabular}[c]{@{}l@{}}1.9 vs \\ \textbf{1.1}\end{tabular}}   & \multicolumn{1}{l|}{\textbf{4E-05}} & 0.025                                                      \\ \cline{2-11} 
                                                     & \begin{tabular}[c]{@{}l@{}}EBO-1 vs \\ DF-PU\end{tabular}                    & \multicolumn{1}{l|}{\begin{tabular}[c]{@{}l@{}}\textbf{1.25} vs \\ 1.75\end{tabular}} & \multicolumn{1}{l|}{\textbf{0.006}} & 0.025                                                      & \multicolumn{1}{l|}{\begin{tabular}[c]{@{}l@{}}\textbf{1.15} vs \\ 1.85\end{tabular}} & \multicolumn{1}{l|}{\textbf{0.001}} & 0.05                                                       & \multicolumn{1}{l|}{\begin{tabular}[c]{@{}l@{}}1.55 vs \\ \textbf{1.45}\end{tabular}} & \multicolumn{1}{l|}{0.701}          & 0.05                                                       \\ \cline{2-11} 
                                                     & \begin{tabular}[c]{@{}l@{}}EBO-1 vs \\ S-EM\end{tabular}                     & \multicolumn{1}{l|}{\begin{tabular}[c]{@{}l@{}}\textbf{1.4} vs \\ 1.6\end{tabular}}   & \multicolumn{1}{l|}{0.498}          & 0.05                                                       & \multicolumn{1}{l|}{\begin{tabular}[c]{@{}l@{}}\textbf{1.15} vs \\ 1.85\end{tabular}} & \multicolumn{1}{l|}{\textbf{1E-04}} & 0.025                                                      & \multicolumn{1}{l|}{\begin{tabular}[c]{@{}l@{}}1.9 vs \\ \textbf{1.1}\end{tabular}}   & \multicolumn{1}{l|}{\textbf{4E-05}} & 0.025                                                      \\ \hline
\end{tabular}
\end{table}

Considering all of the results shown in Table \ref{tab8}, all three Auto-PU systems performed favourably across all comparisons against the baselines (DF-PU and S-EM), outperforming the baselines in every case for F-measure and precision. For recall, in general, the baseline methods performed better, in many cases with statistical significance. 

As shown by Figure \ref{fig3}, although the Auto-PU systems experience a significant performance decline for \(\delta\) = 60\% (as mentioned earlier), S-EM’s performance remains consistent throughout. However, note that the average performance of S-EM is lower than all versions of Auto-PU for all \(\delta\) values, so whilst the performance of Auto-PU is unstable across \(\delta\) values, even the worst average performance of Auto-PU is better than the best average performance of either baseline. 

Considering all results from this Section (6.1), we can conclude that the Auto-PU systems outperform the baselines. The only measure on which the Auto-PU systems are outperformed by the baselines is recall, but this is largely due to their overpredictions of the positive class, which raises recall but consequently decreased precision due to the increase in the number of false positive predictions, leading to a reduced F-measure value overall. 

However, predictive accuracy is not the only metric to consider when evaluating these three Auto-PU systems, especially since none achieved statistically significant superiority when compared against the other two systems. Computational cost is also a factor to consider, due to the very large computational cost generally associated with Auto-ML systems. To that end, the computational runtimes to run a 5-fold cross-validation per dataset of these systems have been measured, running on a 48 core GPU, and are as follows: GA-1: 226.3 minutes, BO-1: 8.4 minutes, EBO-1: 24.2 minutes. Thus BO-1 ran about 27 times faster than GA-1, and almost three times as fast as EBO-1. Therefore, whilst BO-1 may not have come out on top in regard to predictive performance when compared with EBO-1 and GA-1, it still shows a very large improvement over GA-1 in regard to computational efficiency and is likely still a worthwhile option for users, given that it was not outperformed with statistical significance. Similarly, EBO-1 performed almost 10 times faster than GA-1, also representing a very significant improvement over GA-1 in regard to computational cost. 

Since EBO-1 performed overall best against the baselines, alongside the improvements in computational time over GA-1, arguably EBO-1 is the best of the three systems, outperforming both baselines with statistical significance for F-measure and precision for all three \(\delta\)  values, with a good trade-off between GA-1 and BO-1 regarding computational time. 

Regarding the correlation between the percentage of positive instances and F-measure (Table \ref{tab6}), S-EM shows a stronger correlation across all three \(\delta\) values than the Auto-PU systems. DF-PU exhibits a similar correlation to the Auto-PU systems.

\subsection{Results for the Three Auto-PU systems with the extended search space}

Recall that the extended search space allows for PU learning algorithms to be developed utilising the spy approach (unlike the base search space). 

\subsubsection{Comparing the three Auto-PU systems}

Table \ref{tab9} reports the F-measure values achieved by the three Auto-PU systems with the extended search space on each of the 20 datasets, for \(\delta\) = 20\%, 40\%, 60\%; whilst 
Table \ref{tab10} reports the statistical significance results comparing the Auto-PU systems in terms of F-measure, precision and recall. 
Table \ref{tab10} has the same structure as Table \ref{tab5} (see Section 6.1.1). Recall that in each cell comparing the average ranks of a pair of Auto-PU systems, the best (lower) rank is shown in boldface, and if the result is statistically significant (i.e. if the p-value is smaller than the adjusted \(\alpha\)) the p-value is also shown in boldface. 


\begin{table}[]
\caption{F-measure results of the Auto-PU systems utilising the extended search space.}\label{tab9}
\centering
\small
\begin{tabular}{|l|lll|lll|lll|}
\hline
\multirow{2}{*}{Dataset} &
  \multicolumn{3}{c|}{\(\delta\) = 20\%} &
  \multicolumn{3}{c|}{\(\delta\) = 40\%} &
  \multicolumn{3}{c|}{\(\delta\) = 60\%} \\ \cline{2-10} 
 &
  \multicolumn{1}{l|}{EBO-2} &
  \multicolumn{1}{l|}{BO-2} &
  GA-2 &
  \multicolumn{1}{l|}{EBO-2} &
  \multicolumn{1}{l|}{BO-2} &
  GA-2 &
  \multicolumn{1}{l|}{EBO-2} &
  \multicolumn{1}{l|}{BO-2} &
  GA-2 \\ \hline
Alzheimer’s &
  \multicolumn{1}{l|}{0.559} &
  \multicolumn{1}{l|}{0.580} &
  0.548 &
  \multicolumn{1}{l|}{0.597} &
  \multicolumn{1}{l|}{0.603} &
  0.576 &
  \multicolumn{1}{l|}{0.581} &
  \multicolumn{1}{l|}{0.492} &
  0.529 \\ \hline
Autism &
  \multicolumn{1}{l|}{0.964} &
  \multicolumn{1}{l|}{0.963} &
  0.982 &
  \multicolumn{1}{l|}{0.938} &
  \multicolumn{1}{l|}{0.937} &
  0.940 &
  \multicolumn{1}{l|}{0.887} &
  \multicolumn{1}{l|}{0.914} &
  0.927 \\ \hline
Breast cancer Coi. &
  \multicolumn{1}{l|}{0.967} &
  \multicolumn{1}{l|}{0.667} &
  0.711 &
  \multicolumn{1}{l|}{0.952} &
  \multicolumn{1}{l|}{0.618} &
  0.671 &
  \multicolumn{1}{l|}{0.923} &
  \multicolumn{1}{l|}{0.000} &
  0.553 \\ \hline
Breast cancer Wis. &
  \multicolumn{1}{l|}{0.882} &
  \multicolumn{1}{l|}{0.959} &
  0.956 &
  \multicolumn{1}{l|}{0.863} &
  \multicolumn{1}{l|}{0.942} &
  0.936 &
  \multicolumn{1}{l|}{0.839} &
  \multicolumn{1}{l|}{0.889} &
  0.866 \\ \hline
Breast cancer mut. &
  \multicolumn{1}{l|}{0.666} &
  \multicolumn{1}{l|}{0.890} &
  0.896 &
  \multicolumn{1}{l|}{0.655} &
  \multicolumn{1}{l|}{0.853} &
  0.739 &
  \multicolumn{1}{l|}{0.587} &
  \multicolumn{1}{l|}{0.845} &
  0.872 \\ \hline
Cervical cancer &
  \multicolumn{1}{l|}{0.867} &
  \multicolumn{1}{l|}{0.867} &
  0.867 &
  \multicolumn{1}{l|}{0.904} &
  \multicolumn{1}{l|}{0.867} &
  0.839 &
  \multicolumn{1}{l|}{0.516} &
  \multicolumn{1}{l|}{0.839} &
  0.350 \\ \hline
Cirrhosis &
  \multicolumn{1}{l|}{0.506} &
  \multicolumn{1}{l|}{0.497} &
  0.446 &
  \multicolumn{1}{l|}{0.493} &
  \multicolumn{1}{l|}{0.515} &
  0.397 &
  \multicolumn{1}{l|}{0.322} &
  \multicolumn{1}{l|}{0.472} &
  0.204 \\ \hline
Dermatology &
  \multicolumn{1}{l|}{0.857} &
  \multicolumn{1}{l|}{0.876} &
  0.901 &
  \multicolumn{1}{l|}{0.891} &
  \multicolumn{1}{l|}{0.841} &
  0.896 &
  \multicolumn{1}{l|}{0.750} &
  \multicolumn{1}{l|}{0.795} &
  0.692 \\ \hline
PI Diabetes &
  \multicolumn{1}{l|}{0.668} &
  \multicolumn{1}{l|}{0.653} &
  0.642 &
  \multicolumn{1}{l|}{0.665} &
  \multicolumn{1}{l|}{0.648} &
  0.646 &
  \multicolumn{1}{l|}{0.607} &
  \multicolumn{1}{l|}{0.615} &
  0.634 \\ \hline
ES Diabetes &
  \multicolumn{1}{l|}{0.957} &
  \multicolumn{1}{l|}{0.954} &
  0.978 &
  \multicolumn{1}{l|}{0.905} &
  \multicolumn{1}{l|}{0.891} &
  0.887 &
  \multicolumn{1}{l|}{0.915} &
  \multicolumn{1}{l|}{0.912} &
  0.894 \\ \hline
Heart Disease &
  \multicolumn{1}{l|}{0.826} &
  \multicolumn{1}{l|}{0.844} &
  0.836 &
  \multicolumn{1}{l|}{0.804} &
  \multicolumn{1}{l|}{0.817} &
  0.780 &
  \multicolumn{1}{l|}{0.747} &
  \multicolumn{1}{l|}{0.805} &
  0.786 \\ \hline
Heart Failure &
  \multicolumn{1}{l|}{0.741} &
  \multicolumn{1}{l|}{0.757} &
  0.751 &
  \multicolumn{1}{l|}{0.656} &
  \multicolumn{1}{l|}{0.652} &
  0.670 &
  \multicolumn{1}{l|}{0.514} &
  \multicolumn{1}{l|}{0.600} &
  0.671 \\ \hline
Hepatitis C &
  \multicolumn{1}{l|}{0.907} &
  \multicolumn{1}{l|}{0.964} &
  0.944 &
  \multicolumn{1}{l|}{0.907} &
  \multicolumn{1}{l|}{0.761} &
  0.863 &
  \multicolumn{1}{l|}{0.689} &
  \multicolumn{1}{l|}{0.612} &
  0.610 \\ \hline
Kidney Disease &
  \multicolumn{1}{l|}{0.911} &
  \multicolumn{1}{l|}{0.976} &
  0.925 &
  \multicolumn{1}{l|}{0.897} &
  \multicolumn{1}{l|}{0.976} &
  0.951 &
  \multicolumn{1}{l|}{0.656} &
  \multicolumn{1}{l|}{0.789} &
  0.806 \\ \hline
Liver Disease &
  \multicolumn{1}{l|}{0.832} &
  \multicolumn{1}{l|}{0.822} &
  0.831 &
  \multicolumn{1}{l|}{0.800} &
  \multicolumn{1}{l|}{0.815} &
  0.817 &
  \multicolumn{1}{l|}{0.748} &
  \multicolumn{1}{l|}{0.722} &
  0.748 \\ \hline
Maternal Risk &
  \multicolumn{1}{l|}{0.854} &
  \multicolumn{1}{l|}{0.847} &
  0.862 &
  \multicolumn{1}{l|}{0.810} &
  \multicolumn{1}{l|}{0.786} &
  0.813 &
  \multicolumn{1}{l|}{0.731} &
  \multicolumn{1}{l|}{0.729} &
  0.738 \\ \hline
Parkinsons &
  \multicolumn{1}{l|}{0.914} &
  \multicolumn{1}{l|}{0.936} &
  0.935 &
  \multicolumn{1}{l|}{0.850} &
  \multicolumn{1}{l|}{0.837} &
  0.843 &
  \multicolumn{1}{l|}{0.720} &
  \multicolumn{1}{l|}{0.800} &
  0.792 \\ \hline
Parkinsons Biom. &
  \multicolumn{1}{l|}{0.259} &
  \multicolumn{1}{l|}{0.286} &
  0.282 &
  \multicolumn{1}{l|}{0.276} &
  \multicolumn{1}{l|}{0.000} &
  0.259 &
  \multicolumn{1}{l|}{0.203} &
  \multicolumn{1}{l|}{0.000} &
  0.280 \\ \hline
Spine &
  \multicolumn{1}{l|}{0.942} &
  \multicolumn{1}{l|}{0.941} &
  0.923 &
  \multicolumn{1}{l|}{0.920} &
  \multicolumn{1}{l|}{0.936} &
  0.917 &
  \multicolumn{1}{l|}{0.802} &
  \multicolumn{1}{l|}{0.700} &
  0.761 \\ \hline
Stroke &
  \multicolumn{1}{l|}{0.232} &
  \multicolumn{1}{l|}{0.256} &
  0.241 &
  \multicolumn{1}{l|}{0.225} &
  \multicolumn{1}{l|}{0.255} &
  0.239 &
  \multicolumn{1}{l|}{0.201} &
  \multicolumn{1}{l|}{0.233} &
  0.243 \\ \hline
\end{tabular}
\end{table}

The results shown in Table  \ref{tab10} are somewhat mixed for F-measure, with GA-2 performing best overall for \(\delta\) = 20\% and 60\%, but worst for \(\delta\) = 40\%, where EBO-2 performs best. These results are consistent for precision, with GA-2 again performing best for \(\delta\) = 20\% and 60\%, but EBO-2 performing best for \(\delta\) = 40\%. In fact, EBO-2 outperforms GA-2 with statistical significance for \(\delta\) = 40\%, the only statistically significant result in the table. For recall, BO-2 achieved slightly better results than the others, but with no statistically significant values. Overall, conclusions regarding the best of the three systems are hard to draw from these results, as the results were somewhat mixed and statistical significance was not achieved in most cases. However, GA-2 achieved overall the best ranks, but it was outperformed in one case by EBO-2 with statistical significance for \(\delta\) = 40\%.  

\begin{table}[htbp]
\caption{Results of Wilcoxon signed-rank tests with Holm correction for multiple hypothesis when comparing each pair of Auto-PU systems (with the extended search space) regarding F-measure, Precision and Recall, for the 3 \(\delta\) values.}
\label{tab10}
\centering
\small
\setlength{\tabcolsep}{3pt}
\begin{tabular}{|l|l|lll|lll|lll|}
\hline
\multicolumn{1}{|c|}{\multirow{2}{*}{\(\delta\) (\%)}} & \multicolumn{1}{c|}{\multirow{2}{*}{\begin{tabular}[c]{@{}c@{}}Compared \\ systems\end{tabular}}} & \multicolumn{3}{c|}{F-measure}                                                                                                                                                                  & \multicolumn{3}{c|}{Precision}                                                                                                                                                                & \multicolumn{3}{c|}{Recall}                                                                                                                                                                   \\ \cline{3-11} 
\multicolumn{1}{|c|}{}                                 & \multicolumn{1}{c|}{}                                                                             & \multicolumn{1}{c|}{\begin{tabular}[c]{@{}c@{}}Avg.   \\ rank\end{tabular}}  & \multicolumn{1}{c|}{p-value} & \multicolumn{1}{c|}{\begin{tabular}[c]{@{}c@{}}adj.   \\ \(\alpha\)\end{tabular}} & \multicolumn{1}{c|}{\begin{tabular}[c]{@{}c@{}}Avg.   \\ rank\end{tabular}}  & \multicolumn{1}{c|}{p-value} & \multicolumn{1}{c|}{\begin{tabular}[c]{@{}c@{}}adj.\\  \(\alpha\)\end{tabular}} & \multicolumn{1}{c|}{\begin{tabular}[c]{@{}c@{}}Avg.   \\ rank\end{tabular}}  & \multicolumn{1}{c|}{p-value} & \multicolumn{1}{c|}{\begin{tabular}[c]{@{}c@{}}adj.\\  \(\alpha\)\end{tabular}} \\ \hline
\multirow{3}{*}{20\%}                                  & \begin{tabular}[c]{@{}l@{}}GA-2 vs \\ BO-2\end{tabular}                                           & \multicolumn{1}{l|}{\begin{tabular}[c]{@{}l@{}}1.5 vs \\ 1.5\end{tabular}}   & \multicolumn{1}{l|}{0.514}   & 0.017                                                                             & \multicolumn{1}{l|}{\begin{tabular}[c]{@{}l@{}}1.5 vs \\ 1.5\end{tabular}}   & \multicolumn{1}{l|}{0.647}   & 0.05                                                                            & \multicolumn{1}{l|}{\begin{tabular}[c]{@{}l@{}}1.58 vs \\ 1.42\end{tabular}} & \multicolumn{1}{l|}{0.376}   & 0.017                                                                           \\ \cline{2-11} 
                                                       & \begin{tabular}[c]{@{}l@{}}GA-2 vs \\ EBO-2\end{tabular}                                          & \multicolumn{1}{l|}{\begin{tabular}[c]{@{}l@{}}\textbf{1.38} vs \\ 1.62\end{tabular}} & \multicolumn{1}{l|}{0.658}   & 0.025                                                                             & \multicolumn{1}{l|}{\begin{tabular}[c]{@{}l@{}}\textbf{1.45} vs \\ 1.55\end{tabular}} & \multicolumn{1}{l|}{0.632}   & 0.025                                                                           & \multicolumn{1}{l|}{\begin{tabular}[c]{@{}l@{}}\textbf{1.42} vs \\ 1.58\end{tabular}} & \multicolumn{1}{l|}{0.758}   & 0.05                                                                            \\ \cline{2-11} 
                                                       & \begin{tabular}[c]{@{}l@{}}BO-2 vs \\ EBO-2\end{tabular}                                          & \multicolumn{1}{l|}{\begin{tabular}[c]{@{}l@{}}1.62 vs \\ \textbf{1.38}\end{tabular}} & \multicolumn{1}{l|}{0.825}   & 0.05                                                                              & \multicolumn{1}{l|}{\begin{tabular}[c]{@{}l@{}}1.52 vs \\ \textbf{1.48}\end{tabular}} & \multicolumn{1}{l|}{0.368}   & 0.017                                                                           & \multicolumn{1}{l|}{\begin{tabular}[c]{@{}l@{}}\textbf{1.45} vs \\ 1.55\end{tabular}} & \multicolumn{1}{l|}{0.396}   & 0.025                                                                           \\ \hline
\multirow{3}{*}{40\%}                                  & \begin{tabular}[c]{@{}l@{}}GA-2 vs \\ BO-2\end{tabular}                                           & \multicolumn{1}{l|}{\begin{tabular}[c]{@{}l@{}}1.55 vs \\ \textbf{1.45}\end{tabular}} & \multicolumn{1}{l|}{0.729}   & 0.05                                                                              & \multicolumn{1}{l|}{\begin{tabular}[c]{@{}l@{}}1.52 vs \\ \textbf{1.48}\end{tabular}} & \multicolumn{1}{l|}{0.387}   & 0.05                                                                            & \multicolumn{1}{l|}{\begin{tabular}[c]{@{}l@{}}\textbf{1.38} vs \\ 1.62\end{tabular}} & \multicolumn{1}{l|}{0.520}   & 0.05                                                                            \\ \cline{2-11} 
                                                       & \begin{tabular}[c]{@{}l@{}}GA-2 vs \\ EBO-2\end{tabular}                                          & \multicolumn{1}{l|}{\begin{tabular}[c]{@{}l@{}}1.65 vs \\ \textbf{1.35}\end{tabular}} & \multicolumn{1}{l|}{0.039}   & 0.017                                                                             & \multicolumn{1}{l|}{\begin{tabular}[c]{@{}l@{}}1.7 vs \\ \textbf{1.3}\end{tabular}}   & \multicolumn{1}{l|}{\textbf{0.013}}   & 0.017                                                                           & \multicolumn{1}{l|}{\begin{tabular}[c]{@{}l@{}}1.62 vs \\ \textbf{1.38}\end{tabular}} & \multicolumn{1}{l|}{0.356}   & 0.017                                                                           \\ \cline{2-11} 
                                                       & \begin{tabular}[c]{@{}l@{}}BO-2 vs \\ EBO-2\end{tabular}                                          & \multicolumn{1}{l|}{\begin{tabular}[c]{@{}l@{}}1.52 vs \\ \textbf{1.48}\end{tabular}} & \multicolumn{1}{l|}{0.220}   & 0.025                                                                             & \multicolumn{1}{l|}{\begin{tabular}[c]{@{}l@{}}1.62 vs \\ \textbf{1.38}\end{tabular}} & \multicolumn{1}{l|}{0.297}   & 0.025                                                                           & \multicolumn{1}{l|}{\begin{tabular}[c]{@{}l@{}}1.52 vs \\ \textbf{1.48}\end{tabular}} & \multicolumn{1}{l|}{0.376}   & 0.025                                                                           \\ \hline
\multirow{3}{*}{60\%}                                  & \begin{tabular}[c]{@{}l@{}}GA-2 vs \\ BO-2\end{tabular}                                           & \multicolumn{1}{l|}{\begin{tabular}[c]{@{}l@{}}\textbf{1.4} vs \\ 1.6\end{tabular}}   & \multicolumn{1}{l|}{0.388}   & 0.017                                                                             & \multicolumn{1}{l|}{\begin{tabular}[c]{@{}l@{}}\textbf{1.42} vs \\ 1.58\end{tabular}} & \multicolumn{1}{l|}{0.387}   & 0.025                                                                           & \multicolumn{1}{l|}{\begin{tabular}[c]{@{}l@{}}\textbf{1.45} vs \\ 1.55\end{tabular}} & \multicolumn{1}{l|}{0.784}   & 0.05                                                                            \\ \cline{2-11} 
                                                       & \begin{tabular}[c]{@{}l@{}}GA-2 vs \\ EBO-2\end{tabular}                                          & \multicolumn{1}{l|}{\begin{tabular}[c]{@{}l@{}}1.5 vs \\ 1.5\end{tabular}}   & \multicolumn{1}{l|}{0.756}   & 0.025                                                                             & \multicolumn{1}{l|}{\begin{tabular}[c]{@{}l@{}}1.52 vs \\ \textbf{1.48}\end{tabular}} & \multicolumn{1}{l|}{0.702}   & 0.05                                                                            & \multicolumn{1}{l|}{\begin{tabular}[c]{@{}l@{}}\textbf{1.45} vs \\ 1.55\end{tabular}} & \multicolumn{1}{l|}{0.622}   & 0.025                                                                           \\ \cline{2-11} 
                                                       & \begin{tabular}[c]{@{}l@{}}BO-2 vs \\ EBO-2\end{tabular}                                          & \multicolumn{1}{l|}{\begin{tabular}[c]{@{}l@{}}\textbf{1.4} vs \\ 1.6\end{tabular}}   & \multicolumn{1}{l|}{0.956}   & 0.05                                                                              & \multicolumn{1}{l|}{\begin{tabular}[c]{@{}l@{}}1.6 vs \\ \textbf{1.4}\end{tabular}}   & \multicolumn{1}{l|}{0.231}   & 0.017                                                                           & \multicolumn{1}{l|}{\begin{tabular}[c]{@{}l@{}}\textbf{1.42} vs \\ 1.58\end{tabular}} & \multicolumn{1}{l|}{0.587}   & 0.017                                                                           \\ \hline
\end{tabular}
\end{table}

\begin{figure}[htbp]
    \centering
    \includegraphics[width=0.6\textwidth]{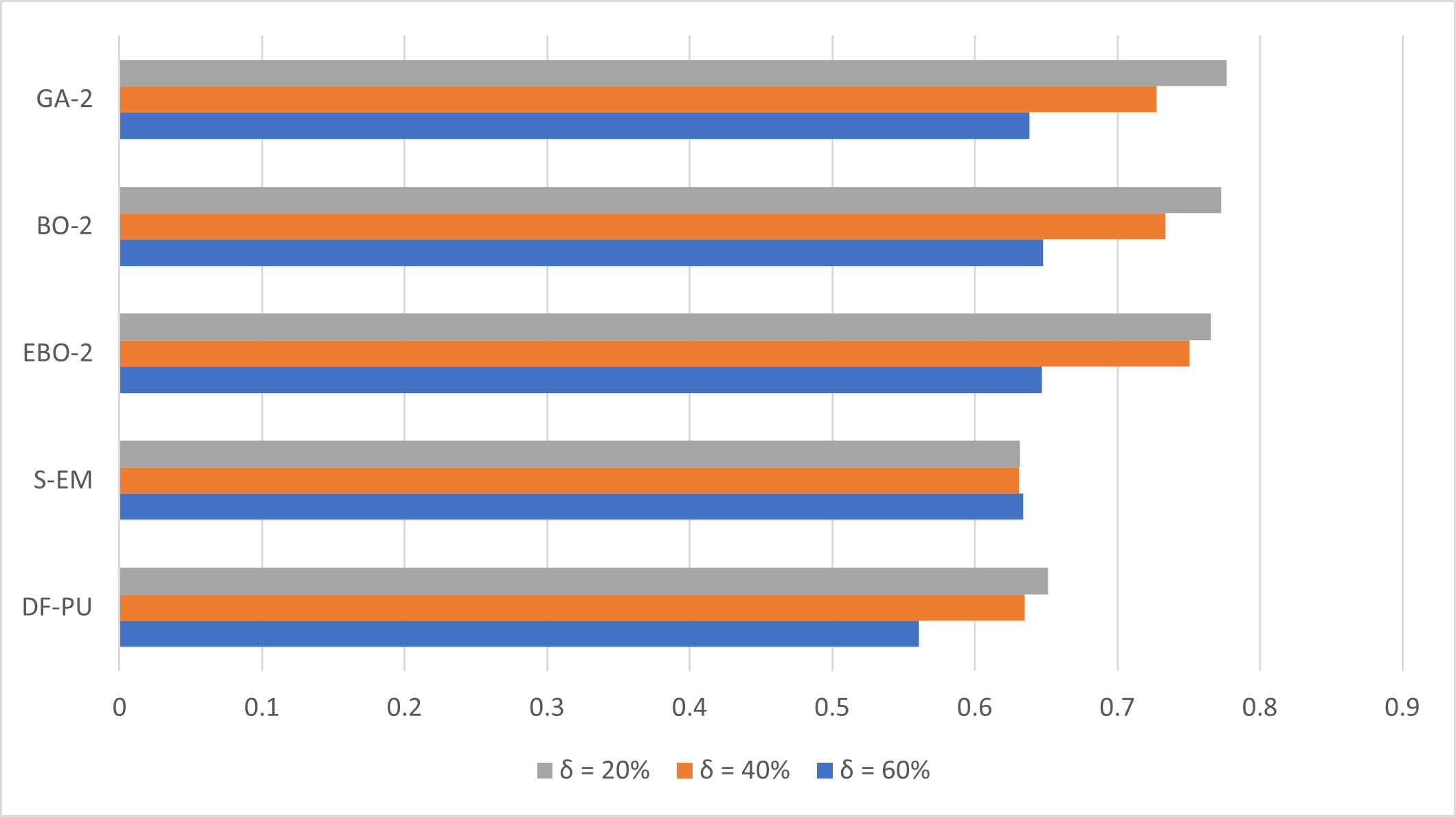}
    \caption{Average F-measure results comparison for three versions of Auto-PU utilising the base search space and the PU learning baselines across the three values of \(\delta\).}
    \label{fig4}
\end{figure}

Figure \ref{fig4} shows how the F-measure values achieved by the three Auto-PU systems and the two baseline PU learning methods (DF-PU and S-EM) vary across the three \(\delta\) values. The results of the baseline methods will be discussed later. Focusing for now on the results for GA-2, BO-2 and EBO-2 in this figure, it is clear that, for all these systems, their F-measure values decrease with an increase in the \(\delta\) value, which is consistent with the results obtained with the base search space (Figure \ref{fig3}). As mentioned earlier, this is not surprising, considering that \(\delta\) represents the percentage of positive instances hidden in the unlabelled set; and intuitively, the larger this percentage, the more difficult the PU learning problem is. Again (like for the results in Figure \ref{fig3}), the decrease in F-measure values was particularly substantial when \(\delta\) increased from 40\% to 60\%, for all the three Auto-PU systems.

To further analyse the results, Table \ref{tab11} shows the values of Pearson’s linear correlation coefficient between the F-measure values achieved by each Auto-PU system or baseline PU learning method and percentages of positive examples in the original dataset, for each \(\delta\) value. Table \ref{tab11} shows the same trends of positive correlations as Table \ref{tab6}; i.e., as the percentage of positive instances increases, the F-measure trends upwards. This, again, holds true for all versions of Auto-PU and for all \(\delta\) values, further implying that PU methods generally perform better when the percentage of positive instances is higher, as expected. The strongest correlation in Table \ref{tab11} is for GA-2 for \(\delta\) = 60\%. Interestingly, BO-2 shows a weak positive correlation for this \(\delta\) value. Overall, BO-2 shows a lower positive correlation than the other two methods, albeit by a very small margin for \(\delta\) = 20\% and \(\delta\) = 40\%.

\begin{table}[htbp]
\caption{Linear (Pearson’s) correlation coefficient value between the F-measure and the percentage of positive examples in the original dataset (before hiding some positive examples in the unlabelled set) for each combination of an Auto-PU system (with extended search space) or PU method and a \(\delta\) value}\label{tab11}
\centering
\small
\begin{tabular}{|l|l|l|l|}
\hline
\multicolumn{1}{|c|}{Method} & \multicolumn{1}{c|}{\(\delta\)   = 20\%} & \multicolumn{1}{c|}{\(\delta\) = 40\%} & \multicolumn{1}{c|}{\(\delta\)  = 60\%} \\ \hline
EBO-2                        & 0.363                                    & 0.361                                  & 0.447                                   \\ \hline
BO-2                         & 0.339                                    & 0.348                                  & 0.225                                   \\ \hline
GA-2                         & 0.340                                    & 0.357                                  & 0.580                                   \\ \hline
DF-PU                        & 0.381                                    & 0.504                                  & 0.445                                   \\ \hline
S-EM                         & 0.690                                    & 0.631                                  & 0.686                                   \\ \hline
\end{tabular}
\end{table}


\subsubsection{Comparing the Auto-PU Systems against two baseline PU methods}

\begin{table}[htbp]
\caption{Results of Wilcoxon signed-rank tests with Holm correction for multiple hypothesis when comparing GA-2, BO-2 and EBO-2 against the two baseline PU methods regarding F-measure, Precision and Recall, for the 3 \(\delta\) values}
\label{tab12}
\centering
\small
\setlength{\tabcolsep}{3pt}
\begin{tabular}{|c|l|lll|lll|lll|}
\hline
\multirow{2}{*}{\(\delta\)   (\%)} & \multicolumn{1}{c|}{\multirow{2}{*}{\begin{tabular}[c]{@{}c@{}}Compared\\ methods\end{tabular}}} & \multicolumn{3}{c|}{F-measure}                                                                                                                                                 & \multicolumn{3}{c|}{Precision}                                                                                                                                                 & \multicolumn{3}{c|}{Recall}                                                                                                                                                    \\ \cline{3-11} 
                                   & \multicolumn{1}{c|}{}                                                                            & \multicolumn{1}{l|}{Ranking}                                                & \multicolumn{1}{l|}{P-value}        & \begin{tabular}[c]{@{}l@{}}Adj. \\ \(\alpha\)\end{tabular} & \multicolumn{1}{l|}{Ranking}                                                & \multicolumn{1}{l|}{P-value}        & \begin{tabular}[c]{@{}l@{}}Adj. \\ \(\alpha\)\end{tabular} & \multicolumn{1}{l|}{Ranking}                                                & \multicolumn{1}{l|}{P-value}        & \begin{tabular}[c]{@{}l@{}}Adj. \\ \(\alpha\)\end{tabular} \\ \hline
\multirow{6}{*}{20}                & \begin{tabular}[c]{@{}l@{}}GA-2 vs\\ DF-PU\end{tabular}                                          & \multicolumn{1}{l|}{\begin{tabular}[c]{@{}l@{}}\textbf{1.1} vs\\ 1.9\end{tabular}}   & \multicolumn{1}{l|}{\textbf{3E-05}} & 0.025                                                      & \multicolumn{1}{l|}{\begin{tabular}[c]{@{}l@{}}\textbf{1.05} vs\\ 1.95\end{tabular}} & \multicolumn{1}{l|}{\textbf{1E-04}} & 0.025                                                      & \multicolumn{1}{l|}{\begin{tabular}[c]{@{}l@{}}\textbf{1.45} vs\\ 1.55\end{tabular}} & \multicolumn{1}{l|}{0.812}          & 0.05                                                       \\ \cline{2-11} 
                                   & \begin{tabular}[c]{@{}l@{}}GA-2 vs\\ S-EM\end{tabular}                                           & \multicolumn{1}{l|}{\begin{tabular}[c]{@{}l@{}}\textbf{1.2} vs\\ 1.8\end{tabular}}   & \multicolumn{1}{l|}{\textbf{0.002}} & 0.05                                                       & \multicolumn{1}{l|}{\begin{tabular}[c]{@{}l@{}}\textbf{1.12} vs\\ 1.88\end{tabular}} & \multicolumn{1}{l|}{\textbf{0.001}} & 0.05                                                       & \multicolumn{1}{l|}{\begin{tabular}[c]{@{}l@{}}1.8 vs\\ \textbf{1.2}\end{tabular}}   & \multicolumn{1}{l|}{\textbf{0.009}} & 0.025                                                      \\ \cline{2-11} 
                                   & \begin{tabular}[c]{@{}l@{}}BO-2 vs\\ DF-PU\end{tabular}                                          & \multicolumn{1}{l|}{\begin{tabular}[c]{@{}l@{}}\textbf{1.1} vs\\ 1.9\end{tabular}}   & \multicolumn{1}{l|}{\textbf{4E-05}} & 0.025                                                      & \multicolumn{1}{l|}{\begin{tabular}[c]{@{}l@{}}\textbf{1.05} vs\\ 1.95\end{tabular}} & \multicolumn{1}{l|}{\textbf{2E-04}} & 0.025                                                      & \multicolumn{1}{l|}{\begin{tabular}[c]{@{}l@{}}\textbf{1.45} vs\\ 1.55\end{tabular}} & \multicolumn{1}{l|}{0.784}          & 0.05                                                       \\ \cline{2-11} 
                                   & \begin{tabular}[c]{@{}l@{}}BO-2 vs\\ S-EM\end{tabular}                                           & \multicolumn{1}{l|}{\begin{tabular}[c]{@{}l@{}}\textbf{1.25} vs\\ 1.75\end{tabular}} & \multicolumn{1}{l|}{\textbf{0.003}} & 0.05                                                       & \multicolumn{1}{l|}{\begin{tabular}[c]{@{}l@{}}\textbf{1.08} vs\\ 1.92\end{tabular}} & \multicolumn{1}{l|}{\textbf{0.001}} & 0.05                                                       & \multicolumn{1}{l|}{\begin{tabular}[c]{@{}l@{}}1.85 vs\\ \textbf{1.15}\end{tabular}} & \multicolumn{1}{l|}{\textbf{0.017}} & 0.025                                                      \\ \cline{2-11} 
                                   & \begin{tabular}[c]{@{}l@{}}EBO-2 vs\\ DF-PU\end{tabular}                                         & \multicolumn{1}{l|}{\begin{tabular}[c]{@{}l@{}}1.2 vs\\ 1.8\end{tabular}}   & \multicolumn{1}{l|}{\textbf{2E-04}} & 0.025                                                      & \multicolumn{1}{l|}{\begin{tabular}[c]{@{}l@{}}\textbf{1.1} vs\\ 1.9\end{tabular}}   & \multicolumn{1}{l|}{\textbf{5E-05}} & 0.025                                                      & \multicolumn{1}{l|}{\begin{tabular}[c]{@{}l@{}}1.52 vs\\ \textbf{1.48}\end{tabular}} & \multicolumn{1}{l|}{0.968}          & 0.05                                                       \\ \cline{2-11} 
                                   & \begin{tabular}[c]{@{}l@{}}EBO-2 vs\\ S-EM\end{tabular}                                          & \multicolumn{1}{l|}{\begin{tabular}[c]{@{}l@{}}\textbf{1.3} vs\\ 1.7\end{tabular}}   & \multicolumn{1}{l|}{\textbf{0.001}} & 0.05                                                       & \multicolumn{1}{l|}{\begin{tabular}[c]{@{}l@{}}\textbf{1.05} vs\\ 1.95\end{tabular}} & \multicolumn{1}{l|}{\textbf{8E-05}} & 0.05                                                       & \multicolumn{1}{l|}{\begin{tabular}[c]{@{}l@{}}1.8 vs\\ \textbf{1.2}\end{tabular}}   & \multicolumn{1}{l|}{\textbf{0.012}} & 0.025                                                      \\ \hline
\multirow{6}{*}{40}                & \begin{tabular}[c]{@{}l@{}}GA-2 vs\\ DF-PU\end{tabular}                                          & \multicolumn{1}{l|}{\begin{tabular}[c]{@{}l@{}}\textbf{1.3} vs\\ 1.7\end{tabular}}   & \multicolumn{1}{l|}{\textbf{0.027}} & 0.05                                                       & \multicolumn{1}{l|}{\begin{tabular}[c]{@{}l@{}}\textbf{1.12} vs\\ 1.88\end{tabular}} & \multicolumn{1}{l|}{\textbf{0.003}} & 0.025                                                      & \multicolumn{1}{l|}{\begin{tabular}[c]{@{}l@{}}1.58 vs\\ \textbf{1.42}\end{tabular}} & \multicolumn{1}{l|}{0.586}          & 0.05                                                       \\ \cline{2-11} 
                                   & \begin{tabular}[c]{@{}l@{}}GA-2 vs\\ S-EM\end{tabular}                                           & \multicolumn{1}{l|}{\begin{tabular}[c]{@{}l@{}}\textbf{1.2} vs\\ 1.8\end{tabular}}   & \multicolumn{1}{l|}{\textbf{0.011}} & 0.025                                                      & \multicolumn{1}{l|}{\begin{tabular}[c]{@{}l@{}}\textbf{1.18} vs\\ 1.82\end{tabular}} & \multicolumn{1}{l|}{\textbf{0.005}} & 0.05                                                       & \multicolumn{1}{l|}{\begin{tabular}[c]{@{}l@{}}1.85 vs\\ \textbf{1.15}\end{tabular}} & \multicolumn{1}{l|}{\textbf{0.001}} & 0.025                                                      \\ \cline{2-11} 
                                   & \begin{tabular}[c]{@{}l@{}}BO-2 vs\\ DF-PU\end{tabular}                                          & \multicolumn{1}{l|}{\begin{tabular}[c]{@{}l@{}}\textbf{1.25} vs\\ 1.75\end{tabular}} & \multicolumn{1}{l|}{\textbf{0.007}} & 0.025                                                      & \multicolumn{1}{l|}{\begin{tabular}[c]{@{}l@{}}\textbf{1.12} vs\\ 1.88\end{tabular}} & \multicolumn{1}{l|}{\textbf{0.001}} & 0.025                                                      & \multicolumn{1}{l|}{\begin{tabular}[c]{@{}l@{}}1.55 vs\\ \textbf{1.45}\end{tabular}} & \multicolumn{1}{l|}{0.701}          & 0.05                                                       \\ \cline{2-11} 
                                   & \begin{tabular}[c]{@{}l@{}}BO-2 vs\\ S-EM\end{tabular}                                           & \multicolumn{1}{l|}{\begin{tabular}[c]{@{}l@{}}\textbf{1.25} vs\\ 1.75\end{tabular}} & \multicolumn{1}{l|}{\textbf{0.008}} & 0.05                                                       & \multicolumn{1}{l|}{\begin{tabular}[c]{@{}l@{}}\textbf{1.18} vs\\ 1.82\end{tabular}} & \multicolumn{1}{l|}{\textbf{0.003}} & 0.05                                                       & \multicolumn{1}{l|}{\begin{tabular}[c]{@{}l@{}}1.8 vs\\ \textbf{1.2}\end{tabular}}   & \multicolumn{1}{l|}{\textbf{0.007}} & 0.025                                                      \\ \cline{2-11} 
                                   & \begin{tabular}[c]{@{}l@{}}EBO-2 vs\\ DF-PU\end{tabular}                                         & \multicolumn{1}{l|}{\begin{tabular}[c]{@{}l@{}}\textbf{1.2} vs\\ 1.8\end{tabular}}   & \multicolumn{1}{l|}{\textbf{0.001}} & 0.025                                                      & \multicolumn{1}{l|}{\begin{tabular}[c]{@{}l@{}}\textbf{1.15} vs\\ 1.85\end{tabular}} & \multicolumn{1}{l|}{\textbf{2E-04}} & 0.025                                                      & \multicolumn{1}{l|}{\begin{tabular}[c]{@{}l@{}}1.52 vs\\ \textbf{1.48}\end{tabular}} & \multicolumn{1}{l|}{0.717}          & 0.05                                                       \\ \cline{2-11} 
                                   & \begin{tabular}[c]{@{}l@{}}EBO-2 vs\\ S-EM\end{tabular}                                          & \multicolumn{1}{l|}{\begin{tabular}[c]{@{}l@{}}\textbf{1.2} vs\\ 1.8\end{tabular}}   & \multicolumn{1}{l|}{\textbf{0.001}} & 0.05                                                       & \multicolumn{1}{l|}{\begin{tabular}[c]{@{}l@{}}\textbf{1.15} vs\\ 1.85\end{tabular}} & \multicolumn{1}{l|}{\textbf{5E-04}} & 0.05                                                       & \multicolumn{1}{l|}{\begin{tabular}[c]{@{}l@{}}1.75 vs\\ \textbf{1.25}\end{tabular}} & \multicolumn{1}{l|}{\textbf{0.024}} & 0.025                                                      \\ \hline
\multirow{6}{*}{60}                & \begin{tabular}[c]{@{}l@{}}GA-2 vs\\ DF-PU\end{tabular}                                          & \multicolumn{1}{l|}{\begin{tabular}[c]{@{}l@{}}\textbf{1.3} vs\\ 1.7\end{tabular}}   & \multicolumn{1}{l|}{0.048}          & 0.025                                                      & \multicolumn{1}{l|}{\begin{tabular}[c]{@{}l@{}}\textbf{1.2} vs\\ 1.8\end{tabular}}   & \multicolumn{1}{l|}{\textbf{0.015}} & 0.05                                                       & \multicolumn{1}{l|}{\begin{tabular}[c]{@{}l@{}}1.52 vs\\ \textbf{1.48}\end{tabular}} & \multicolumn{1}{l|}{0.658}          & 0.05                                                       \\ \cline{2-11} 
                                   & \begin{tabular}[c]{@{}l@{}}GA-2 vs\\ S-EM\end{tabular}                                           & \multicolumn{1}{l|}{\begin{tabular}[c]{@{}l@{}}\textbf{1.4} vs\\ 1.6\end{tabular}}   & \multicolumn{1}{l|}{0.546}          & 0.05                                                       & \multicolumn{1}{l|}{\begin{tabular}[c]{@{}l@{}}\textbf{1.12} vs\\ 1.88\end{tabular}} & \multicolumn{1}{l|}{\textbf{0.006}} & 0.025                                                      & \multicolumn{1}{l|}{\begin{tabular}[c]{@{}l@{}}1.85 vs\\ \textbf{1.15}\end{tabular}} & \multicolumn{1}{l|}{\textbf{8E-05}} & 0.025                                                      \\ \cline{2-11} 
                                   & \begin{tabular}[c]{@{}l@{}}BO-2 vs\\ DF-PU\end{tabular}                                          & \multicolumn{1}{l|}{\begin{tabular}[c]{@{}l@{}}\textbf{1.35} vs\\ 1.65\end{tabular}} & \multicolumn{1}{l|}{0.070}          & 0.025                                                      & \multicolumn{1}{l|}{\begin{tabular}[c]{@{}l@{}}\textbf{1.15} vs\\ 1.85\end{tabular}} & \multicolumn{1}{l|}{\textbf{0.005}} & 0.05                                                       & \multicolumn{1}{l|}{\begin{tabular}[c]{@{}l@{}}1.5 vs\\ 1.5\end{tabular}}   & \multicolumn{1}{l|}{0.571}          & 0.05                                                       \\ \cline{2-11} 
                                   & \begin{tabular}[c]{@{}l@{}}BO-2 vs\\ S-EM\end{tabular}                                           & \multicolumn{1}{l|}{\begin{tabular}[c]{@{}l@{}}1.5 vs\\ 1.5\end{tabular}}   & \multicolumn{1}{l|}{0.571}          & 0.05                                                       & \multicolumn{1}{l|}{\begin{tabular}[c]{@{}l@{}}\textbf{1.12 }vs\\ 1.88\end{tabular}} & \multicolumn{1}{l|}{\textbf{3E-04}} & 0.025                                                      & \multicolumn{1}{l|}{\begin{tabular}[c]{@{}l@{}}1.9 vs\\ \textbf{1.1}\end{tabular}}   & \multicolumn{1}{l|}{\textbf{1E-05}} & 0.025                                                      \\ \cline{2-11} 
                                   & \begin{tabular}[c]{@{}l@{}}EBO-2 vs\\ DF-PU\end{tabular}                                         & \multicolumn{1}{l|}{\begin{tabular}[c]{@{}l@{}}\textbf{1.3} vs\\ 1.7\end{tabular}}   & \multicolumn{1}{l|}{0.027}          & 0.025                                                      & \multicolumn{1}{l|}{\begin{tabular}[c]{@{}l@{}}\textbf{1.4} vs\\ 1.6\end{tabular}}   & \multicolumn{1}{l|}{0.143}          & 0.05                                                       & \multicolumn{1}{l|}{\begin{tabular}[c]{@{}l@{}}\textbf{1.3} vs\\ 1.7\end{tabular}}   & \multicolumn{1}{l|}{0.294}          & 0.05                                                       \\ \cline{2-11} 
                                   & \begin{tabular}[c]{@{}l@{}}EBO-2 vs\\ S-EM\end{tabular}                                          & \multicolumn{1}{l|}{\begin{tabular}[c]{@{}l@{}}1.5 vs\\ 1.5\end{tabular}}   & \multicolumn{1}{l|}{0.841}          & 0.05                                                       & \multicolumn{1}{l|}{\begin{tabular}[c]{@{}l@{}}\textbf{1.28} vs\\ 1.72\end{tabular}} & \multicolumn{1}{l|}{\textbf{0.016}} & 0.025                                                      & \multicolumn{1}{l|}{\begin{tabular}[c]{@{}l@{}}1.8 vs\\ \textbf{1.2}\end{tabular}}   & \multicolumn{1}{l|}{\textbf{0.001}} & 0.025                                                      \\ \hline
\end{tabular}
\end{table}

As the full F-measure results for the Auto-PU systems and the baseline PU methods have already been given (in Tables \ref{tab4} and \ref{tab7} respectively), these results will not be repeated here. However, Table \ref{tab12} shows the statistical significance results of comparing each of the three Auto-PU systems against each of the two baseline PU methods. Table \ref{tab12} has the same structure as the previously described Table \ref{tab8}.

As discussed in Section 6.1, computational expense is also a factor to consider when utilising Auto-ML systems. The average runtimes of the systems to complete a 5-fold cross-validation per dataset, running on a 48 core GPU, were as follows: GA-2: 223.2 minutes, BO-2: 9.8 minutes, EBO-2: 20.2 minutes. Thus BO-2 ran about 23 times faster than GA-2, and just over twice as fast as EBO-2. EBO-2 ran about 11 times faster than GA-2. 

Overall, considering all the results from this Section 6.2, the three Auto-PU systems using the extended search space have clearly outperformed the baseline PU learning methods. 

\subsection{Analysis of the Most Frequently Selected Hyperparameter Values}

In this section we review and analyse the most frequently selected hyperparameter values of the optimised PU learning algorithms output by the Auto-PU systems.  These results are shown in Tables \ref{tab13} - \ref{tab15}. Each table shows the results for one of the three Auto-PU systems, and for each hyperparameter shown in the first column, there are two rows, reporting the results for each of the two versions of the system (with base or extended search space); with the next columns showing the most selected value of the hyperparameter, its selection frequency, the baseline frequency and the difference between the selection and baseline frequencies. The selection frequency is the percentage of times that hyperparameter value was selected out of all runs of that version of the Auto-PU system, over all 10 iterations of the external cross-validation procedure for all 20 datasets and all three values of \(\delta\) (i.e. the selection \% out of 600 cases). The baseline frequency is the expected selection frequency of a hyperparameter value if all values of that hyperparameter were randomly selected for use in a PU learning algorithm. I.e., it is calculated by simply dividing 1 (one) by the number of candidate values for that hyperparameter. Note that in Tables \ref{tab13} - \ref{tab15} there are no results for the Spy-related hyperparameters (“N/A” = not applicable) in the rows for the base search space because the Spy method is used only in the extended search space.

\begin{table}[htbp]
\caption{Hyperparameter values most frequently selected by GA-Auto-PU}\label{tab13}
\centering
\small
\begin{tabular}{|l|l|l|ll|ll|ll|}
\hline
\multicolumn{1}{|c|}{\textbf{Hyperparameter}} & \multicolumn{1}{c|}{\textbf{\begin{tabular}[c]{@{}c@{}}Search \\ space\end{tabular}}} & \multicolumn{1}{c|}{\textbf{\begin{tabular}[c]{@{}c@{}}Most selected \\ value\end{tabular}}} & \multicolumn{2}{c|}{\textbf{\begin{tabular}[c]{@{}c@{}}Selection \\ Freq. (\%)\end{tabular}}} & \multicolumn{2}{c|}{\textbf{\begin{tabular}[c]{@{}c@{}}Baseline \\ Freq. (\%)\end{tabular}}} & \multicolumn{2}{c|}{\textbf{\begin{tabular}[c]{@{}c@{}}Diff.\\ (\%)\end{tabular}}} \\ \hline
\multirow{2}{*}{Phase 1A Iteration Count}     & base                                                                                  & 1                                                                                            & \multicolumn{2}{l|}{19.67}                                                                    & \multicolumn{2}{l|}{10.00}                                                                   & \multicolumn{2}{l|}{9.67}                                                          \\ \cline{2-9} 
                                              & extended                                                                              & 1                                                                                            & \multicolumn{2}{l|}{26.67}                                                                    & \multicolumn{2}{l|}{10.00}                                                                   & \multicolumn{2}{l|}{16.67}                                                         \\ \hline
\multirow{2}{*}{Phase 1A RN Threshold}        & base                                                                                  & 0.3                                                                                          & \multicolumn{2}{l|}{15.00}                                                                    & \multicolumn{2}{l|}{10.00}                                                                   & \multicolumn{2}{l|}{5.00}                                                          \\ \cline{2-9} 
                                              & extended                                                                              & 0.3                                                                                          & \multicolumn{2}{l|}{20.33}                                                                    & \multicolumn{2}{l|}{10.00}                                                                   & \multicolumn{2}{l|}{10.33}                                                         \\ \hline
\multirow{2}{*}{Phase 1A Classifier}          & base                                                                                  & Gaussian NB                                                                                  & \multicolumn{2}{l|}{14.00}                                                                    & \multicolumn{2}{l|}{5.56}                                                                    & \multicolumn{2}{l|}{8.44}                                                          \\ \cline{2-9} 
                                              & extended                                                                              & Random forest                                                                                & \multicolumn{2}{l|}{12.67}                                                                    & \multicolumn{2}{l|}{5.56}                                                                    & \multicolumn{2}{l|}{7.11}                                                          \\ \hline
\multirow{2}{*}{Phase 1B Flag}                & base                                                                                  & False                                                                                        & \multicolumn{2}{l|}{52.33}                                                                    & \multicolumn{2}{l|}{50.00}                                                                   & \multicolumn{2}{l|}{2.33}                                                          \\ \cline{2-9} 
                                              & extended                                                                              & False                                                                                        & \multicolumn{2}{l|}{66.00}                                                                    & \multicolumn{2}{l|}{50.00}                                                                   & \multicolumn{2}{l|}{16.00}                                                         \\ \hline
\multirow{2}{*}{Phase 1B RN Threshold}        & base                                                                                  & 0.15                                                                                         & \multicolumn{2}{l|}{19.00}                                                                    & \multicolumn{2}{l|}{10.00}                                                                   & \multicolumn{2}{l|}{9.00}                                                          \\ \cline{2-9} 
                                              & extended                                                                              & 0.25                                                                                         & \multicolumn{2}{l|}{15.33}                                                                    & \multicolumn{2}{l|}{10.00}                                                                   & \multicolumn{2}{l|}{5.33}                                                          \\ \hline
\multirow{2}{*}{Phase 1B Classifier}          & base                                                                                  & Deep forest                                                                                  & \multicolumn{2}{l|}{11.00}                                                                    & \multicolumn{2}{l|}{5.56}                                                                    & \multicolumn{2}{l|}{5.44}                                                          \\ \cline{2-9} 
                                              & extended                                                                              & SVM                                                                                          & \multicolumn{2}{l|}{10.67}                                                                    & \multicolumn{2}{l|}{5.56}                                                                    & \multicolumn{2}{l|}{5.11}                                                          \\ \hline
\multirow{2}{*}{Spy rate}                     & base                                                                                  & N/A                                                                                          & \multicolumn{2}{l|}{N/A}                                                                      & \multicolumn{2}{l|}{N/A}                                                                     & \multicolumn{2}{l|}{N/A}                                                           \\ \cline{2-9} 
                                              & extended                                                                              & 0.1                                                                                          & \multicolumn{2}{l|}{18.67}                                                                    & \multicolumn{2}{l|}{14.29}                                                                   & \multicolumn{2}{l|}{4.38}                                                          \\ \hline
\multirow{2}{*}{Spy tolerance}                & base                                                                                  & N/A                                                                                          & \multicolumn{2}{l|}{N/A}                                                                      & \multicolumn{2}{l|}{N/A}                                                                     & \multicolumn{2}{l|}{N/A}                                                           \\ \cline{2-9} 
                                              & extended                                                                              & 0.06                                                                                         & \multicolumn{2}{l|}{18.47}                                                                    & \multicolumn{2}{l|}{9.09}                                                                    & \multicolumn{2}{l|}{9.38}                                                          \\ \hline
\multirow{2}{*}{Spy flag}                     & base                                                                                  & N/A                                                                                          & \multicolumn{2}{l|}{N/A}                                                                      & \multicolumn{2}{l|}{N/A}                                                                     & \multicolumn{2}{l|}{N/A}                                                           \\ \cline{2-9} 
                                              & extended                                                                              & False                                                                                        & \multicolumn{2}{l|}{73.33}                                                                    & \multicolumn{2}{l|}{50.00}                                                                   & \multicolumn{2}{l|}{23.33}                                                         \\ \hline
\multirow{2}{*}{Phase 2 Classifier}           & base                                                                                  & LDA                                                                                          & \multicolumn{2}{l|}{21.00}                                                                    & \multicolumn{2}{l|}{5.56}                                                                    & \multicolumn{2}{l|}{15.44}                                                         \\ \cline{2-9} 
                                              & extended                                                                              & Deep forest                                                                                  & \multicolumn{2}{l|}{10.00}                                                                    & \multicolumn{2}{l|}{5.56}                                                                    & \multicolumn{2}{l|}{4.44}                                                          \\ \hline
\end{tabular}
\end{table}

\begin{table}[]
\caption{Hyperparameter values most frequently selected by BO-Auto-PU}\label{tab14}
\centering
\small
\begin{tabular}{|l|l|l|ll|ll|ll|}
\hline
\textbf{Hyperparameter}                   & \textbf{\begin{tabular}[c]{@{}l@{}}Search \\ space\end{tabular}} & \textbf{\begin{tabular}[c]{@{}l@{}}Most selected \\ value\end{tabular}} & \multicolumn{2}{l|}{\textbf{\begin{tabular}[c]{@{}l@{}}Selection\\ Freq. (\%)\end{tabular}}} & \multicolumn{2}{l|}{\textbf{\begin{tabular}[c]{@{}l@{}}Baseline \\ Freq. (\%)\end{tabular}}} & \multicolumn{2}{l|}{\textbf{\begin{tabular}[c]{@{}l@{}}Diff.\\ (\%)\end{tabular}}} \\ \hline
\multirow{2}{*}{Phase 1A Iteration Count} & base                                                             & 2                                                                       & \multicolumn{2}{l|}{19.00}                                                                   & \multicolumn{2}{l|}{10.00}                                                                   & \multicolumn{2}{l|}{9.00}                                                          \\ \cline{2-9} 
                                          & extended                                                         & 2                                                                       & \multicolumn{2}{l|}{21.00}                                                                   & \multicolumn{2}{l|}{10.00}                                                                   & \multicolumn{2}{l|}{11.00}                                                         \\ \hline
\multirow{2}{*}{Phase 1A RN Threshold}    & base                                                             & 0.05                                                                    & \multicolumn{2}{l|}{14.33}                                                                   & \multicolumn{2}{l|}{10.00}                                                                   & \multicolumn{2}{l|}{4.33}                                                          \\ \cline{2-9} 
                                          & extended                                                         & 0.25                                                                    & \multicolumn{2}{l|}{13.00}                                                                   & \multicolumn{2}{l|}{10.00}                                                                   & \multicolumn{2}{l|}{3.00}                                                          \\ \hline
\multirow{2}{*}{Phase 1A Classifier}      & base                                                             & Bernoulli NB                                                            & \multicolumn{2}{l|}{8.67}                                                                    & \multicolumn{2}{l|}{5.56}                                                                    & \multicolumn{2}{l|}{3.11}                                                          \\ \cline{2-9} 
                                          & extended                                                         & Logistic reg.                                                           & \multicolumn{2}{l|}{8.67}                                                                    & \multicolumn{2}{l|}{5.56}                                                                    & \multicolumn{2}{l|}{3.11}                                                          \\ \hline
\multirow{2}{*}{Phase 1B Flag}            & base                                                             & True                                                                    & \multicolumn{2}{l|}{52.67}                                                                   & \multicolumn{2}{l|}{50.00}                                                                   & \multicolumn{2}{l|}{2.67}                                                          \\ \cline{2-9} 
                                          & extended                                                         & True                                                                    & \multicolumn{2}{l|}{50.67}                                                                   & \multicolumn{2}{l|}{50.00}                                                                   & \multicolumn{2}{l|}{0.67}                                                          \\ \hline
\multirow{2}{*}{Phase 1B RN Threshold}    & base                                                             & 0.2                                                                     & \multicolumn{2}{l|}{14.00}                                                                   & \multicolumn{2}{l|}{10.00}                                                                   & \multicolumn{2}{l|}{4.00}                                                          \\ \cline{2-9} 
                                          & extended                                                         & 0.2                                                                     & \multicolumn{2}{l|}{14.67}                                                                   & \multicolumn{2}{l|}{10.00}                                                                   & \multicolumn{2}{l|}{4.67}                                                          \\ \hline
\multirow{2}{*}{Phase 1B Classifier}      & base                                                             & HGBoost                                                                 & \multicolumn{2}{l|}{8.00}                                                                    & \multicolumn{2}{l|}{5.56}                                                                    & \multicolumn{2}{l|}{2.44}                                                          \\ \cline{2-9} 
                                          & extended                                                         & Bagging clas.                                                           & \multicolumn{2}{l|}{7.67}                                                                    & \multicolumn{2}{l|}{5.56}                                                                    & \multicolumn{2}{l|}{2.11}                                                          \\ \hline
\multirow{2}{*}{Spy rate}                 & base                                                             & N/A                                                                     & \multicolumn{2}{l|}{N/A}                                                                     & \multicolumn{2}{l|}{N/A}                                                                     & \multicolumn{2}{l|}{N/A}                                                           \\ \cline{2-9} 
                                          & extended                                                         & 0.3                                                                     & \multicolumn{2}{l|}{18.00}                                                                   & \multicolumn{2}{l|}{14.29}                                                                   & \multicolumn{2}{l|}{3.71}                                                          \\ \hline
\multirow{2}{*}{Spy tolerance}            & base                                                             & N/A                                                                     & \multicolumn{2}{l|}{N/A}                                                                     & \multicolumn{2}{l|}{N/A}                                                                     & \multicolumn{2}{l|}{N/A}                                                           \\ \cline{2-9} 
                                          & extended                                                         & 0.08                                                                    & \multicolumn{2}{l|}{12.18}                                                                   & \multicolumn{2}{l|}{9.09}                                                                    & \multicolumn{2}{l|}{3.09}                                                          \\ \hline
\multirow{2}{*}{Spy flag}                 & base                                                             & N/A                                                                     & \multicolumn{2}{l|}{N/A}                                                                     & \multicolumn{2}{l|}{N/A}                                                                     & \multicolumn{2}{l|}{N/A}                                                           \\ \cline{2-9} 
                                          & extended                                                         & False                                                                   & \multicolumn{2}{l|}{74.00}                                                                   & \multicolumn{2}{l|}{50.00}                                                                   & \multicolumn{2}{l|}{24.00}                                                         \\ \hline
\multirow{2}{*}{Phase 2 Classifier}       & base                                                             & LDA                                                                     & \multicolumn{2}{l|}{32.67}                                                                   & \multicolumn{2}{l|}{5.56}                                                                    & \multicolumn{2}{l|}{27.11}                                                         \\ \cline{2-9} 
                                          & extended                                                         & LDA                                                                     & \multicolumn{2}{l|}{51.67}                                                                   & \multicolumn{2}{l|}{5.56}                                                                    & \multicolumn{2}{l|}{46.11}                                                         \\ \hline
\end{tabular}
\end{table}

\begin{table}[]
\caption{Hyperparameter values most frequently selected by EBO-Auto-PU}\label{tab15}
\centering
\small
\begin{tabular}{|l|l|l|ll|ll|ll|}
\hline
\multicolumn{1}{|c|}{\textbf{Hyperparameter}} & \multicolumn{1}{c|}{\textbf{\begin{tabular}[c]{@{}c@{}}Search \\ space\end{tabular}}} & \multicolumn{1}{c|}{\textbf{\begin{tabular}[c]{@{}c@{}}Most selected \\ value\end{tabular}}} & \multicolumn{2}{c|}{\textbf{\begin{tabular}[c]{@{}c@{}}Selection \\ Freq. (\%)\end{tabular}}} & \multicolumn{2}{c|}{\textbf{\begin{tabular}[c]{@{}c@{}}Baseline\\ Freq. (\%)\end{tabular}}} & \multicolumn{2}{c|}{\textbf{\begin{tabular}[c]{@{}c@{}}Diff.\\ (\%)\end{tabular}}} \\ \hline
\multirow{2}{*}{Phase 1A Iteration Count}     & base                                                                                  & 2                                                                                            & \multicolumn{2}{l|}{19.67}                                                                    & \multicolumn{2}{l|}{10.00}                                                                  & \multicolumn{2}{l|}{9.67}                                                          \\ \cline{2-9} 
                                              & extended                                                                              & 2                                                                                            & \multicolumn{2}{l|}{22.33}                                                                    & \multicolumn{2}{l|}{10.00}                                                                  & \multicolumn{2}{l|}{12.33}                                                         \\ \hline
\multirow{2}{*}{Phase 1A RN Threshold}        & base                                                                                  & 0.4                                                                                          & \multicolumn{2}{l|}{13.67}                                                                    & \multicolumn{2}{l|}{10.00}                                                                  & \multicolumn{2}{l|}{3.67}                                                          \\ \cline{2-9} 
                                              & extended                                                                              & 0.15                                                                                         & \multicolumn{2}{l|}{18.33}                                                                    & \multicolumn{2}{l|}{10.00}                                                                  & \multicolumn{2}{l|}{8.33}                                                          \\ \hline
\multirow{2}{*}{Phase 1A Classifier}          & base                                                                                  & Logistic reg.                                                                                & \multicolumn{2}{l|}{14.00}                                                                    & \multicolumn{2}{l|}{5.56}                                                                   & \multicolumn{2}{l|}{8.44}                                                          \\ \cline{2-9} 
                                              & extended                                                                              & LDA                                                                                          & \multicolumn{2}{l|}{10.33}                                                                    & \multicolumn{2}{l|}{5.56}                                                                   & \multicolumn{2}{l|}{4.77}                                                          \\ \hline
\multirow{2}{*}{Phase 1B Flag}                & base                                                                                  & True                                                                                         & \multicolumn{2}{l|}{59.67}                                                                    & \multicolumn{2}{l|}{50.00}                                                                  & \multicolumn{2}{l|}{9.67}                                                          \\ \cline{2-9} 
                                              & extended                                                                              & True                                                                                         & \multicolumn{2}{l|}{56.67}                                                                    & \multicolumn{2}{l|}{50.00}                                                                  & \multicolumn{2}{l|}{6.67}                                                          \\ \hline
\multirow{2}{*}{Phase 1B RN Threshold}        & base                                                                                  & 0.4                                                                                          & \multicolumn{2}{l|}{18.00}                                                                    & \multicolumn{2}{l|}{10.00}                                                                  & \multicolumn{2}{l|}{8.00}                                                          \\ \cline{2-9} 
                                              & extended                                                                              & 0.2                                                                                          & \multicolumn{2}{l|}{14.67}                                                                    & \multicolumn{2}{l|}{10.00}                                                                  & \multicolumn{2}{l|}{4.67}                                                          \\ \hline
\multirow{2}{*}{Phase 1B Classifier}          & base                                                                                  & SVM                                                                                          & \multicolumn{2}{l|}{11.67}                                                                    & \multicolumn{2}{l|}{5.56}                                                                   & \multicolumn{2}{l|}{6.11}                                                          \\ \cline{2-9} 
                                              & extended                                                                              & Logistic reg.                                                                                & \multicolumn{2}{l|}{9.00}                                                                     & \multicolumn{2}{l|}{5.56}                                                                   & \multicolumn{2}{l|}{3.44}                                                          \\ \hline
\multirow{2}{*}{Spy rate}                     & base                                                                                  & N/A                                                                                          & \multicolumn{2}{l|}{N/A}                                                                      & \multicolumn{2}{l|}{N/A}                                                                    & \multicolumn{2}{l|}{N/A}                                                           \\ \cline{2-9} 
                                              & extended                                                                              & 0.1                                                                                          & \multicolumn{2}{l|}{24.00}                                                                    & \multicolumn{2}{l|}{5.56}                                                                   & \multicolumn{2}{l|}{18.44}                                                         \\ \hline
\multirow{2}{*}{Spy tolerance}                & base                                                                                  & N/A                                                                                          & \multicolumn{2}{l|}{N/A}                                                                      & \multicolumn{2}{l|}{N/A}                                                                    & \multicolumn{2}{l|}{N/A}                                                           \\ \cline{2-9} 
                                              & extended                                                                              & 0.01                                                                                         & \multicolumn{2}{l|}{13.88}                                                                    & \multicolumn{2}{l|}{10.00}                                                                  & \multicolumn{2}{l|}{3.88}                                                          \\ \hline
\multirow{2}{*}{Spy flag}                     & base                                                                                  & N/A                                                                                          & \multicolumn{2}{l|}{N/A}                                                                      & \multicolumn{2}{l|}{N/A}                                                                    & \multicolumn{2}{l|}{N/A}                                                           \\ \cline{2-9} 
                                              & extended                                                                              & False                                                                                        & \multicolumn{2}{l|}{65.67}                                                                    & \multicolumn{2}{l|}{50.00}                                                                  & \multicolumn{2}{l|}{15.67}                                                         \\ \hline
\multirow{2}{*}{Phase 2 Classifier}           & base                                                                                  & Random forest                                                                                & \multicolumn{2}{l|}{9.33}                                                                     & \multicolumn{2}{l|}{5.5\%}                                                                  & \multicolumn{2}{l|}{3.77}                                                          \\ \cline{2-9} 
                                              & extended                                                                              & Deep forest                                                                                  & \multicolumn{2}{l|}{10.67}                                                                    & \multicolumn{2}{l|}{5.56}                                                                   & \multicolumn{2}{l|}{5.11}                                                          \\ \hline
\end{tabular}
\end{table}

Regarding the Phase 1A Classifier, all Auto-PU systems had a clear preference for simple linear classifiers, with Gaussian naïve Bayes (Table \ref{tab13}), Bernoulli naïve Bayes (Table \ref{tab14}), logistic regression (Tables \ref{tab14} and \ref{tab15}), and linear discriminant analysis (LDA) (Table \ref{tab15}). These classifiers adhere to the assumptions of separability and smoothness (see Section 2.1), key assumptions for the two-step framework. The exception is random forest, the most frequently selected Phase 1A classifier by GA-2 (Table \ref{tab13}). As random forest is an ensemble classifier, it is neither simple nor linear. However, random forest is a powerful classifier, so it is unsurprising that it was frequently selected. There is little cohesion in the Phase 1B classifier, and as such no clear conclusions can be drawn for that hyperparameter. 

There are 2 reoccurring classifiers selected as the Phase 2 classifier for all Auto-PU systems, LDA (Tables \ref{tab13} and \ref{tab14}) and the deep forest classifier (Tables \ref{tab13} and \ref{tab15}). These two classifiers are very different; LDA is a relatively simple linear classifier, whilst deep forest is a complex classifier, drawing inspiration from deep learning and creating an “ensemble of ensembles” \citep{zeng2020predicting}. Deep forest is also the base classifier for one of the baseline approaches used in this work, DF-PU. 

Regarding the Phase 1B flag, almost all results show it set to true at a rate between 50\% and 60\%. As such, the use of Phase 1B is highly dependent on the dataset. 

Regarding the Spy flag parameter for all Auto-PU systems, these results show it set to false far more frequently than set to true in all tables. This is surprising, given the prevalence of the spy method throughout the PU learning literature. These results indicate that it may not be as effective as the frequency of its use would suggest. This further justifies the need for Auto-ML systems such as those proposed throughout this work as, simply based on a literature review of the PU learning literature, one would be forgiven for assuming that spy-based methods are very effective PU learning systems, given the frequency of their use. However, based on these results, it is clear that this is not the case. 

Regarding the Phase 1A iteration count hyperparameter, this hyperparameter splits the unlabelled set into multiple subsets in order to prevent the Phase 1A classifier being overwhelmed by the unlabelled set, given that it is generally the majority class. Therefore, the higher the degree of class imbalance, the higher the Phase 1A iteration count hyperparameter should be. In order to evaluate this, the Pearson’s correlation coefficient was calculated to analyse the correlation between this hyperparameter and the degree of class imbalance, calculated as the percentage of positive instances in the whole dataset. Table \ref{tab16} shows the results of this analysis, with all methods showing a moderate to strong negative correlation, as defined by \citep{schober2018correlation}. Based on these correlations, it can be argued that when assembling a two-step PU learning algorithm, one should consider the class imbalance present when deciding upon the value of the iteration count hyperparameter to apply.

\begin{table}[]
\centering
\caption{Pearson’s correlation coefficient values for Phase 1A iteration count to class imbalance. }\label{tab16}
\centering
\small
\begin{tabular}{|l|l|l|l|}
\hline
\multicolumn{1}{|c|}{\textbf{Method}} & \multicolumn{1}{c|}{\textbf{\(\delta\) = 20\%}} & \multicolumn{1}{c|}{\textbf{\(\delta\) = 40\%}} & \multicolumn{1}{c|}{\textbf{\(\delta\) = 60\%}} \\ \hline
GA-1                                  & -0.646                                          & -0.655                                          & -0.689                                          \\ \hline
GA-2                                  & -0.631                                          & -0.687                                          & -0.723                                          \\ \hline
BO-1                                  & -0.677                                          & -0.700                                          & -0.700                                          \\ \hline
BO-2                                  & -0.641                                          & -0.706                                          & -0.736                                          \\ \hline
EBO-1                                 & -0.656                                          & -0.688                                          & -0.696                                          \\ \hline
EBO-2                                 & -0.680                                          & -0.710                                          & -0.687                                          \\ \hline
\end{tabular}
\end{table}
\section{Conclusions}

This work has introduced two new Auto-ML systems for PU learning (referred to as Auto-PU systems), namely BO-Auto-PU and EBO-Auto-PU, based on Bayesian Optimisation and a new hybrid Evolutionary Bayesian Optimisation method, respectively. These two new Auto-PU systems were compared against GA-Auto-PU,  the only existing Auto-ML system for PU learning, as well as against two baseline PU learning methods, DF-PU (deep forests adapted to PU learning) and S-EM (the Spy method for PU learning). The evaluation of these Auto-PU systems and PU learning methods was based mainly on the F-measure, which is the most popular measure of predictive performance in PU learning \citep{saunders2022evaluating}; but we reported statistical significance results not only for F-measure, but also for precision and recall, for the sake of completeness. The evaluation was performed considering three different versions of each of 20 biomedical datasets, varying the parameter \(\delta\), the percentage of positive training examples hidden in the unlabelled training set (\(\delta\) = 20\%, 40\%, 60\%). Each Auto-PU system has two versions, with the base and extended search spaces, where only the latter includes the Spy method.

The results for the base search space (Section 6.1) have shown that for \(\delta\)=20\% the three Auto-PU systems perform comparably. However, for \(\delta\)=40\% and 60\%, EBO-1 outperforms the other two systems. In general, however, for all results with the base search space, across the three \(\delta\) values, the differences in F-measure, precision and recall values among the three systems are not statistically significant. The comparisons between the Auto-PU systems and the baseline methods have shown all three Auto-PU systems largely outperforming the baselines with statistical significance in most cases regarding F-measure and precision. 

The results for the extended search space (Section 6.2) show that, among the three Auto-PU systems, GA-2 has achieved overall the best ranks for \(\delta\)=20\% and 60\%, whilst EBO-2 has achieved the best ranks for \(\delta\)=40\%. In nearly all cases (combinations of performance measures and \(\delta\) values), there was no statistically significance in the differences of results among GA-2, BO-2 and EBO-2. The exception is that EBO-2 outperformed GA-2 with statistical significance for \(\delta\) = 40\%. All three systems often outperformed the two baseline methods with statistical significance for F-measure and precision.

A correlation analysis has shown that, for the three Auto-PU systems, there was a moderate degree of correlation between the percentage of positive instances in the dataset and the F-measure values achieved by those systems, whilst there was a stronger correlation for S-EM. DF-PU exhibited a similar correlation to the Auto-PU systems.

Regarding computational time, BO-Auto-PU is about 27 times faster than GA-Auto-PU. EBO-Auto-PU sits between the two other Auto-PU systems at approximately 10 times faster than GA-Auto-PU and almost three times as slow as BO-Auto-PU. 

Considering the trade-off between the predictive performance and computational efficiency, it can be argued that EBO-Auto-PU is the best performing system overall, achieving good predictive performance that is often a statistically significant improvement over the baseline PU learning methods, whilst maintaining  a good computational efficiency. 

Future work could involve exploring other search spaces, such as a search space incorporating another PU learning framework such as biased learning \citep{bekker2020learning}. In addition, note that, although each Auto-PU system is optimising the hyperparameters of a PU learning algorithm (in the two-step framework), the hyperparameters of each Auto-PU system itself are currently not optimised. In future work the Auto-PU systems’ hyperparameters could perhaps be optimised too.

\bibliography{sample}
\end{document}